\documentclass[journal]{IEEEtran}

\usepackage{amsmath}
\usepackage{amssymb}
\usepackage{mathtools}
\usepackage{amsthm}

\usepackage{microtype}
\usepackage{verbatim}
\usepackage{cite}
\usepackage{amsfonts}

\usepackage{graphicx}
\usepackage{booktabs} 
\graphicspath{{./figures/}}
\DeclareGraphicsExtensions{.pdf,.jpeg,.png}
\usepackage{units}
\usepackage[caption=false, font=footnotesize]{subfig}
\usepackage{multirow}
\usepackage[inline]{enumitem}
\usepackage{tikz}

\newif\ifasterisk

\usepackage[hyphens]{url}
\usepackage{hyperref}
\usepackage[capitalise]{cleveref}

\usepackage[linesnumbered, ruled,vlined]{algorithm2e}
\SetKwInput{KwInput}{Input}                
\SetKwInput{KwOutput}{Output}              

\usepackage{algcompatible}
\algblockdefx{FORALLP}{ENDFAP}[1]%
  {\textbf{for all }#1 \textbf{do in parallel}}%
  {\textbf{end for}}
\algblockdefx{IF}{END_IF}[1]%
  {\textbf{if }#1 \textbf{then}}%
  {\textbf{end if}}

\DeclareMathOperator*{\argmax}{arg\,max}

\newcommand{\ie}{\textit{i.e.,}\ }
\newcommand{\eg}{\textit{e.g.,}\ } 

\begin{document}

\title{SCOPE: Stochastic Cartographic Occupancy Prediction Engine for Uncertainty-Aware \\ Dynamic Navigation}

\author{Zhanteng Xie,~\IEEEmembership{Member,~IEEE}, and Philip Dames,~\IEEEmembership{Member,~IEEE}
\thanks{*This work was funded by Temple University.}
\thanks{Zhanteng Xie and Philip Dames are with the Department of Mechanical Engineering,
        Temple University, Philadelphia, PA, USA
        {\tt\small \{zhanteng.xie, pdames\}@temple.edu}
        
        Multimedia are available: \url{https://youtu.be/8TtHTtJzuc8}}%
}

\maketitle


\begin{abstract}
This article presents a family of Stochastic Cartographic Occupancy Prediction Engines (SCOPEs) that enable mobile robots to predict the future states of complex dynamic environments. They do this by accounting for the motion of the robot itself, the motion of dynamic objects, and the geometry of static objects in the scene, and they generate a range of possible future states of the environment. These prediction engines are software-optimized for real-time performance for navigation in crowded dynamic scenes, achieving up to 89 times faster inference speed and 8 times less memory usage than other state-of-the-art engines. Three simulated and real-world datasets collected by different robot models are used to demonstrate that these proposed prediction algorithms are able to achieve more accurate and robust stochastic prediction performance than other algorithms. Furthermore, a series of simulation and hardware navigation experiments demonstrate that the proposed predictive uncertainty-aware navigation framework with these stochastic prediction engines is able to improve the safe navigation performance of current state-of-the-art model- and learning-based control policies.
\end{abstract}


\section{Introduction}
\label{sec:introduction}
\begin{figure}[t]
    \centering
    \includegraphics[width=0.45\textwidth]{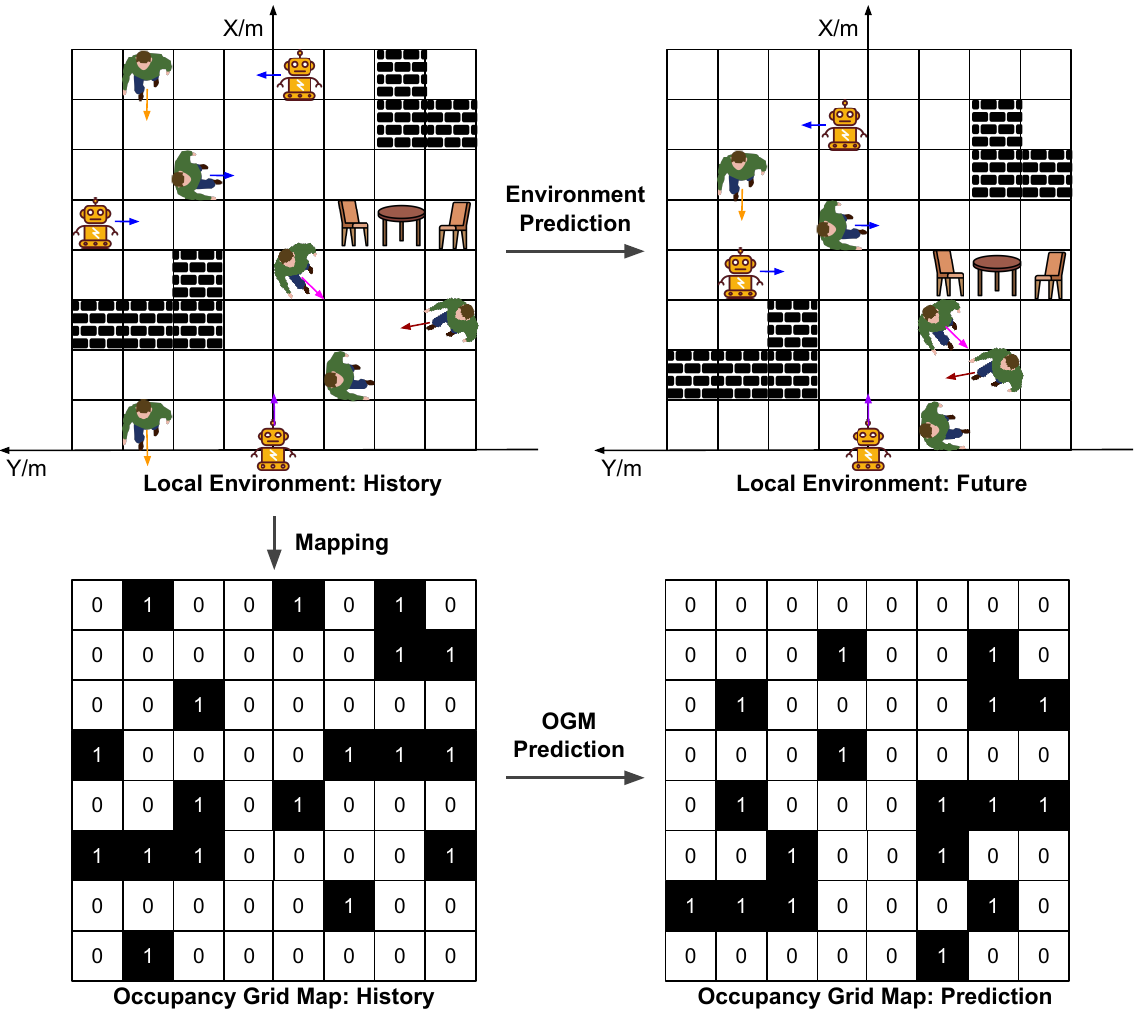}
    \caption{A simple illustration of the occupancy grid map prediction problem. 
    In a complex dynamic environment with many pedestrians, robots, tables, chairs and walls, colored arrows indicate the velocity of each agent.
    }
    \label{fig:ogm_problem}
\end{figure}

\IEEEPARstart{A}{utonomous} mobile robots are beginning to enter people’s lives and are trying to help us provide different last mile delivery services, such as moving goods in warehouses or hospitals and assisting grocery shoppers~\cite{ED2022:online, Keimyung2021:online, Sick2021:online}.
To realize this vision, mobile robots are required to safely and efficiently navigate through complex and dynamic environments filled not only with static obstacles (\eg tables, chairs, and walls) but also with many moving people and/or other mobile robots.
The first prerequisite for robots to navigate and perform tasks is to use their sensors to perceive the surrounding environment.
This work focuses on the next step, which is to accurately and reliably predict how the surrounding environment will change based on these sensor data, as shown in \cref{fig:ogm_problem}.
This will allow a robot to proactively act based on its predictions and the associated uncertainty to avoid potential future collisions, a key part of improving autonomous robot navigation. 
Note that since this general perception-prediction-control navigation framework is a complex and resource-intensive system, it is very important to make these algorithms hardware-friendly (\eg using smaller computational power, memory usage, and storage usage) and run in real-time, especially for mobile robots with limited resources.
A well-performing predictor is useless for practical robotics applications if it consumes a lot of memory and/or cannot run in real-time on a resource-limited robot.

In this article, we propose a family of deep neural network (DNN)-based Stochastic Cartographic Occupancy Prediction Engines (\ie SCOPE, SCOPE++, and SO-SCOPE) for resource-constrained mobile robots to provide stochastic future state predictions and enable uncertainty-aware navigation in crowded dynamic scenes.
This article is an evolution of our previous work \cite{xie2023sogmp}, in which we defined the architecture of the SCOPE and SCOPE++,\footnote{In our previous work \cite{xie2023sogmp} we used the acronym SOGMP (Stochastic Occupancy Grid Map Predictor) instead of SCOPE.} validated their ability to predict occupancy grid maps (OGMs), and demonstrated their utility for robotic navigation.
The primary extensions to~\cite{xie2023sogmp} include: 
\begin{enumerate*}
    \item providing comprehensive OGM prediction performance and resource usage evaluations that include three new state-of-the-art algorithms, 
    \item mathematically modeling the output characteristics of the prior work to understand its performance, particularly around the statistical distribution of the VAE,  
    \item designing a novel software-optimized SO-SCOPE predictor that significantly improves the inference speed and addresses the memory-intensive nature of the previous sampling-based SCOPE and SCOPE++ predictors, 
    \item extending the previous predictive uncertainty-aware navigation framework to enable it to be integrated with resource-intensive learning-based control policies, and
    \item providing additional simulation experiments and real-world experiments to demonstrate the performance of these algorithms. 
\end{enumerate*}

Therefore, integrating our previous work~\cite{xie2023sogmp} and providing more technically improved materials and more complete experiments, this article presents six primary contributions:
\begin{enumerate}
    \item We design an algorithmic pipeline called SCOPE++ that can use a short history of robot odometry and lidar measurements to predict a distribution of potential future robot/environment states. 
    SCOPE++ includes modules to compensate for the ego-motion of the robot, to segment static/dynamic objects in the scene, to predict future scenes using a ConvLSTM network, and to sample other future scenes using a variational autoencoder (VAE).
    
    \item We analyze the running time and memory usage of each module of SCOPE++ to identify computational bottlenecks. 
    Based on this, we compress the VAE by performing an in-depth statistical analysis of its output and by using knowledge distillation techniques.
    The resulting software-optimized SCOPE (SO-SCOPE) achieves slightly better performance while consuming less memory, performing faster inference, and running in real-time with other resource-intensive algorithms on resource-constrained mobile robot hardware.

    \item We validate the ability of our SCOPE predictors (\ie SCOPE++, SCOPE, and SO-SCOPE) to predict OGMs using three OGM datasets (each of which comes from a different robot model) and provide a comprehensive benchmark of prediction performance and resource usage using six state-of-the-art algorithms.
    We find that the SCOPE family achieves smaller absolute errors, higher structural similarity, higher tracking accuracy, and lower computational resource requirements than other state-of-the-art methods (\ie ConvLSTM~\cite{shi2015convolutional}, DeepTracking~\cite{ondruska2016deep}, PhyDNet~\cite{guen2020disentangling}, SAAConvLSTM~\cite{lange2021attention}, TAAConvLSTM~\cite{lange2021attention}, and LOPR~\cite{lange2022lopr}).
    We also perform a detailed analysis of the correctness, diversity, and consistency of the uncertainty estimates from the SCOPE family.
    
    \item We propose a costmap-based predictive uncertainty-aware navigation framework to incorporate OGM prediction and its uncertainty information into current existing navigation control policies to improve their safe navigation performance in crowded dynamic scenes.
    
    \item We validate the navigation performance in simulated 3D environments with varying crowd densities and real-world experiments.
    We find that the predictive uncertainty-aware navigation framework combined with our proposed SCOPE family can improve the navigation performance and safety of extant control policies relative to state-of-the-art solutions, including a model-based controller~\cite{fox1997dynamic}, a supervised learning-based approach~\cite{xie2021towards}, and two deep reinforcement learning (DRL)-based approaches \cite{guldenring2020learning, xie2023drlvo}.
    
    \item We open source the OGM prediction code with the OGM dataset~\cite{sogmp_dataset} (\url{https://github.com/TempleRAIL/scope}) and its predictive uncertainty-aware navigation framework (\url{https://github.com/TempleRAIL/scope_nav}).

\end{enumerate}

\section{Related Works}
\label{sec:related_works}
In this section, we provide a detailed description of prior work on environment prediction, deep neural network compression, and uncertainty-aware navigation. 

\subsection{Environment Prediction}
\label{subsec:environment_prediction}
Environment prediction remains an open problem as the future state of the environment is unknown, complex, and stochastic.
Many interesting works have focused on this prediction problem.
Traditional object detection and tracking methods~\cite{ess2010object, nuss2018random} use multi-stage procedures, hand-designed features, and explicitly detect and track objects.
More recently, deep learning (DL)-based methods that are detection and tracking-free have been able to obtain more accurate predictions \cite{ondruska2016deep, itkina2019dynamic, toyungyernsub2021double, schreiber2019long, schreiber2020motion, lange2021attention}.
Occupancy grid maps (OGMs) are the most common environment representation in these methods.
This transforms the complex environment prediction problem into an OGM prediction problem, outlined in \cref{fig:ogm_problem}.  
Since OGMs can be treated as images (both are 2D arrays of data), the multi-step OGM prediction problem can be thought of as a video prediction task, a well-studied problem in machine learning. 

The most common technique for OGM prediction uses recurrent neural networks (RNNs), which are widely used in video prediction (\eg ConvLSTM~\cite{shi2015convolutional}, PredNet~\cite{lotter2016deep}, and PhyDNet~\cite{guen2020disentangling}).
For example, Ondruska et al.~\cite{ondruska2016deep} first propose an RNN-based deep tracking framework to directly track and predict unoccluded OGM states from raw sensor data.
Itkina et al.~\cite{itkina2019dynamic} directly adapt PredNet~\cite{lotter2016deep} to predict the dynamic OGMs (DOGMas) in urban scenes. 
Following this line of thought, Toyungyernsub et al.~\cite{toyungyernsub2021double} decouple the static and dynamic OGMs and propose a double-prong PredNet to predict occupancy states of the environment. 
Similarly, Schreiber et al.~\cite{schreiber2019long, schreiber2020motion} embed the ConvLSTM units in the U-Net architecture to capture spatio-temporal information of DOGMs and predict them in the stationary vehicle setting.
Lange et al.~\cite{lange2021attention} propose two attention-augmented ConvLSTM networks to capture long-range dependencies and predict future OGMs in the moving vehicle setting.
However, these image-based works only focus on improving network architectures and just treat the OGMs as images, assuming their network architectures can implicitly capture useful information from the kinematics and dynamics behind the environment with sufficient good data.

While pure image-based methods are simple and explicitly ignore the dynamic information of the environment, many other DL-based approaches exploit the ego-motion information and motion flow of the environment to improve the OGM prediction accuracy. 
By using a combination of input placement and recurrent states shifting to compensate for the ego-motion, Schreiber et al.~\cite{schreiber2021dynamic} extend their previous image-based works~\cite{schreiber2019long, schreiber2020motion} to predict DOGMs in moving ego-vehicle scenarios.
Dequaire et al.~\cite{dequaire2018deep} extend the deep tracking framework~\cite{ondruska2016deep} and propose a gated recurrent unit (GRU)-based deep tracking network with a spatial transformer module (STM) for ego-motion compensation to predict multi-steps future states in stationary and moving vehicle settings.
Song et al.~\cite{song20192d} propose a GRU-based LiDAR-FlowNet to estimate the forward and backward motion flow between two consecutive OGMs and predict future OGMs.  
Thomas et al.~\cite{thomas2022learning, thomas2023foreseeable} directly encode spatiotemporal information into the world coordinate frame and propose a 3D-2D feedforward architecture, called DeepSOGM, to predict futures. 
By considering the ego-motion and motion flow together, Mohajerin et al.~\cite{mohajerin2019multi} first use the standard geometric image transformation to compensate for the ego-motion of the vehicle, then propose a ConvLSTM-based difference learning architecture to extract the motion difference between consecutive OGMs, and finally predict multi-step future states.
However, most of these motion-based works are designed for vehicle scenarios, and their network models are memory-intensive and computationally intensive, putting them out of the range of resource-limited mobile robots.
Furthermore, all the above-described works only provide deterministic OGM predictions and cannot estimate the uncertainty of future states, which is a key point in helping robots operate in dynamic hazardous environments.
To address this issue, we propose VAE-based stochastic OGM predictors for resource-constrained robots, namely SCOPE++ and SCOPE, both of which predict a distribution of possible future states of dynamic scenes.

\subsection{Deep Neural Network Optimization}
\label{subsec:deep_neural_network_optimization}
Due to limited computational resources, mobile robots always need hardware-friendly deep learning algorithms.
Many researchers have been working on accelerating deep learning algorithms from hardware and software aspects, or both.
Compared to hardware acceleration, which requires significant redesign and manufacturing changes using specialized hardware architectures such as field programmable gate arrays (FPGAs), neural processing units (NPUs), and custom application-specific integrated circuits (ASICs) (most robots are not equipped with these specialized hardware computing devices), software acceleration offers greater advantages by enabling flexibility and adaptability through modifications to deep neural networks and other software optimizations.

As summarized in surveys~\cite{cheng2018model, mishra2020survey}, the main methods of software acceleration can be divided into three categories: 1) network pruning, 2) weight quantization and sharing, and 3) knowledge distillation.
The key idea of network pruning is to analyze the different components of complex DNNs and remove unimportant components, such as unimportant channels~\cite{he2017channel, howard2017mobilenets}, unimportant kernel filters~\cite{denton2014exploiting, luo2017thinet}, unimportant network connections~\cite{molchanov2017variational, parashar2017scnn}, and even unimportant layers~\cite{he2018multi, tan2019mnasnet}. 
However, network pruning requires expertise and experience to analyze the network model and can easily lead to performance degradation.
To avoid the extra knowledge of pruning, many researchers focus on network model weight quantization and sharing, such as using Huffman coding to quantize weights~\cite{han2015deep}, reduce the number of bits representing model weights~\cite{hubara2016binarized, jacob2018quantization}, and share weights between different network connections or different layers~\cite{han2016eie, georgiev2017low}.
However, weight quantization often requires a good trade-off between performance and bit quantization, and requires computing devices to support its low-precision arithmetic operations.
To keep network model performance and compress neural network, Hinton et al.~\cite{hinton2015distilling} first proposed the knowledge distillation technique, the key idea of which is to train a small and simple ``student'' network to learn the output of a large and complex ``teacher'' network.
Following the basic distillation idea, many improvement variants are proposed to reduce the learning gap between the ``student'' network and the ``teacher'' network such as adding a middle-size network as the ``teaching assistant''~\cite{mirzadeh2020improved, son2021densely}, using early-stopped knowledge distillation ``teacher'' network~\cite{cho2019efficacy}, and incorporating cross-domain distillation training~\cite{yang2020mobileda}.
Although knowledge distillation requires the selection of appropriate ``student'' or ``teacher'' networks, it provides more freedom to choose different compressed network architectures and provides the potential possibility of obtaining compressed ``student'' networks with better generalization capabilities.
Therefore, we chose to combine the knowledge distillation technique with the uncertainty quantification module to software-optimize our proposed VAE-based stochastic OGM predictors and make them hardware-friendly for resource-constrained robots.

\subsection{Uncertainty-Aware Navigation}
\label{subsec:uncertainty_aware_navigation}
As we previously discussed, while most approaches~\cite{fox1997dynamic, guldenring2020learning, xie2021towards, xie2023drlvo} use the past or current environment states (\ie raw sensor data and preprocessed data representations) as perception information for their navigation control policies, many methods~\cite{sathyamoorthy2020densecavoid, chen2020relational, dugas2020navrep, li2020socially, katyal2020intent, liu2023intention} believe that predicted future states can help robots better avoid collisions with pedestrians, and start utilizing the predicted environment states (\eg pedestrian trajectory prediction and latent state prediction) to improve navigation performance through pedestrian crowds.
However, most prediction-based navigation policies assume the outcomes of the interactive environments and their control actions are deterministic.
Due to the stochastic nature of the robot and its surrounding environments, there are many uncertainties in environment prediction and control actions. 
To leverage the uncertainty information in the environmental future states and provide robust and reliable navigation behavior, Kahn et al.~\cite{kahn2017uncertainty} propose a collision prediction network to estimate the future collision uncertainty and incorporate the predicted collision uncertainty into Model Predictive Control (MPC) framework to enable uncertainty-aware navigation behavior. 
Similarly, under the MPC framework, L{\"u}tjens et al.~\cite{lutjens2019safe} use an ensemble of LSTM networks with dropout and bootstrapping to estimate prediction collision probabilities and provide uncertainty-aware navigation around pedestrians.
Tang et al.~\cite{tang2022prediction} use the uncertainty-aware potential field to process prediction uncertainty and incorporate it into the MPC framework to provide predictive uncertainty-aware navigation.
Furthermore, Sekiguchi et al.~\cite{sekiguchi2021uncertainty} convert the uncertainty information of predicted human trajectories into reliability values and incorporate it into a nonlinear MPC framework to realize safe human-followings.
However, all these uncertainty-aware works focus on applying their prediction uncertainty information into the MPC framework and cannot be directly adapted to the currently existing model-based and learning-based control policies.

To easily incorporate prediction uncertainty information into existing model-based and learning-based control policies, some works convert prediction uncertainty information as obstacle costmaps to improve existing path planning modules. 
For example, Georgakis et al.~\cite{georgakis2022uncertainty} propose a UPEN uncertainty-driven planner by combining OGM prediction uncertainty costmaps with an RRT planner~\cite{lavalle2001randomized} and a DD-PPO control policy~\cite{wijmans2019dd}.
However, this UPEN planner only works in static environments. 
To enable dynamic navigation, Thomas et al.~\cite{thomas2023foreseeable} add OGM prediction costmaps into the timed-elastic-band (TEB) planner of the ROS navigation stack framework. 
However, their proposed DeepSOGM predictor is a time-consuming 3D network and only contains prediction information but cannot provide and utilize prediction uncertainty information for dynamic navigation.
To address this issue, we follow the costmap-based navigation framework and propose a hardware-friendly predictive uncertainty-aware navigation framework to incorporate the prediction and uncertainty information into existing model-based and learning-based control policies to improve their safe navigation performance.


\begin{figure*}[t]
    \centering
    \includegraphics[width=0.98\textwidth]{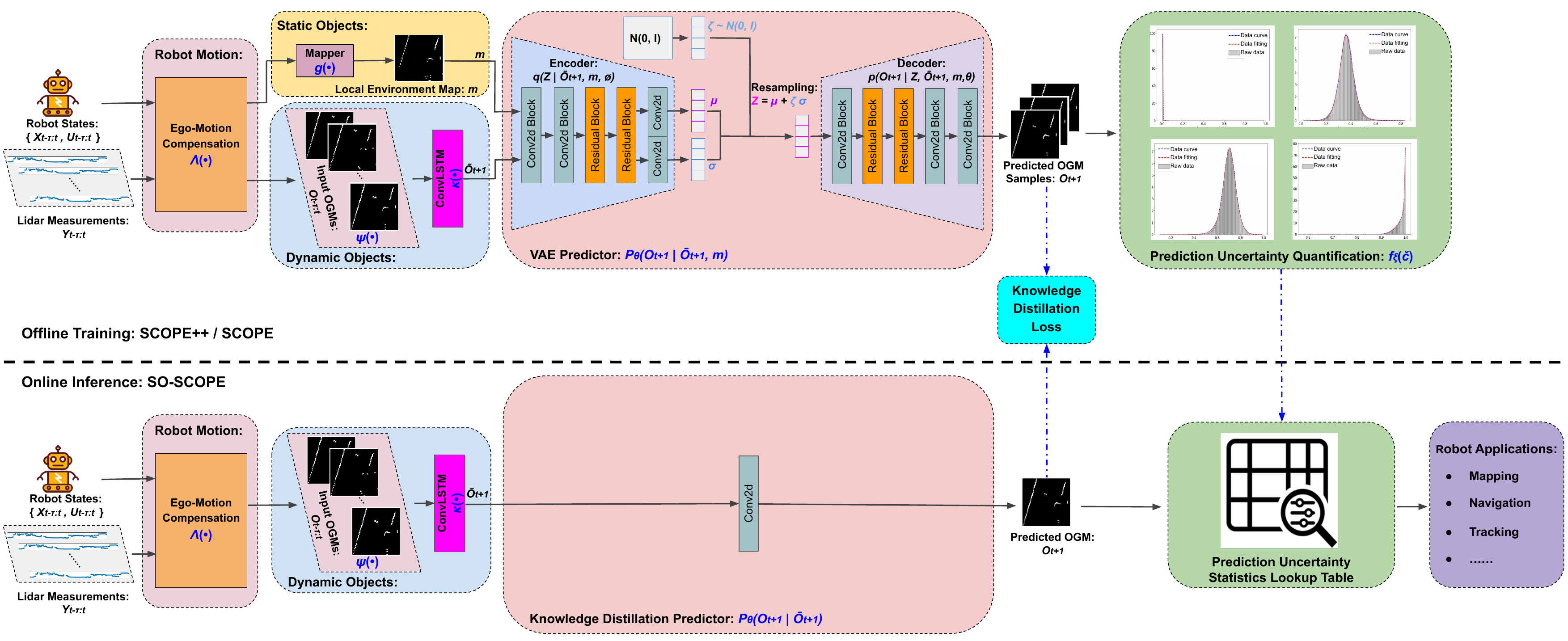}
    \caption{System architectures of the SCOPE++ predictor, SCOPE predictor, and its software-accelerated SO-SCOPE predictor (note that SCOPE omits the Static Objects block compared to SCOPE++).
    The basic process of the SCOPE++ predictor is: 1) based on a history of robot states, the robot transfers the lidar measurement history to the predicted coordinate frame of the robot to compensate for the ego-motion, 2) these compensated lidar measurements are used to generate a local environment map to account for static objects, and a set of OGMs to account for dynamic objects, and 3) the local map of static objects and the predicted OGM of dynamic objects are fed into an variational autoencoder to predict the future OGM. 
    To accelerate the SCOPE++ predictor, we first follow the SCOPE network architecture and replace the VAE network with a single convolutional layer, then use knowledge distillation technology to train the SO-SCOPE network to obtain the prediction information, and finally, we model and quantify the prediction uncertainty of the SCOPE++ to obtain uncertainty statistics and use them to generate uncertainty estimates of SO-SCOPE. 
    }
    \label{fig:scope}
\end{figure*}

\section{Stochastic Cartographic Occupancy Prediction Engine}
\label{sec:stochastic_occupancy_grip_map_predictor}
This section begins by formulating the problem of OGM prediction in dynamic scenes. 
It will then go on to describe our VAE-based predictors (SCOPE series), describing the system architecture as well as how to handle robot motion, pedestrian motion, and static obstacles in dynamic environments.

\subsection{Problem Formulation}
\label{subsec:problem_formulation}
We consider the problem of a mobile robot moving through an environment filled with both static objects (\eg walls) and dynamic objects (\eg people).
As the robot moves, we assume that it can obtain accurate estimates of its relative pose and velocity from odometry sensors or other localization algorithms in a short period of time (on the order of \unit[1]{s}).
We denote the pose and velocity of the robot at time $t$ by $\mathbf{x}_t = [x_t\ y_t\ \theta_t]^T$ and $\mathbf{u}_t = [v_t\ w_t]^T$ respectively. 
We assume the robot is equipped with a 2D lidar sensor.
Let $\mathbf{y}_{t} = [\mathbf{r}_t\ \mathbf{b}_t]^T$ denote the lidar measurements (range $\mathbf{r}$ and bearing $\mathbf{b}$) at time $t$.
We assume the world is 2.5D, as is commonly done in mobile robotics applications. 
Thus, we can represent the environment using a 2D occupancy grid map (OGM).
Let $\mathbf{o}_t$ denote the OGM at time $t$.
We will consider OGMs that are both binary (\ie each cell is either occupied or free) and probabilistic (\ie each cell has a probability of being occupied).

Given a history of $\tau$ lidar measurements $\mathbf{y}_{t-\tau:t}$ and robot states $\{\mathbf{x}_{t-\tau:t}, \mathbf{u}_{t-\tau:t}\}$, we consider the problem of predicting the distribution of future states (\ie OGMs) of the environment $\mathbf{o}_{t+1}$. 
For notational compactness, let $\mathbf{d}_{t-\tau:t} = \{\mathbf{x}_{t-\tau:t}, \mathbf{u}_{t-\tau:t}, \mathbf{y}_{t-\tau:t}\}$ denote the history of data over the given window.
We can formulate this as a prediction model: 
\begin{equation}
    p_{\theta}(\mathbf{o}_{t+1} \mid \mathbf{d}_{t-\tau:t}),
    \label{eq:pf}
\end{equation}
where $\theta$ are the model parameters, and the goal is to find the optimal $\theta^*$ to maximize \eqref{eq:pf}.
Note that this model can also be used in a rollout to predict multiple time steps into the future, \ie using $t+1$ to predict $t+2$, etc.

Note that we set the lidar history to $\tau=10$, the sampling rate is \unit[10]{Hz}, and limit the physical size of OGMs to \unit[{$[0,6.4]$}]{m} along the x-axis (forward) and \unit[{$[-3.2,3.2]$}]{m} along the y-axis (left) in the robot's local coordinate frame.
We use a cell size of \unit[0.1]{m}, resulting in $64 \times 64$ OGMs.
These settings are consistent with other works on mobile robot navigation ~\cite{katyal2019uncertainty}.
All data $\mathbf{o}, \mathbf{u}, \mathbf{y}$ are represented in the robot's \textit{local} coordinate frame.

\subsection{Image-Based Prediction}
\label{subsec:image_based_prediction}
An OGM can be considered a grayscale image, with the probability of occupancy defining the ``color'' in each cell of a regular grid.
Using this interpretation, image/video prediction algorithms could be used to generate the next OGM given a short sequence of past OGMs~\cite{itkina2019dynamic, schreiber2019long, schreiber2020motion}.
The prediction model~\eqref{eq:pf} of image-based approaches is rewritten as
\begin{subequations}
\begin{align}
    \mathbf{o}_{t+1}^* = \argmax \, p_{\theta}(\mathbf{o}_{t+1} \mid \mathbf{d}_{t-\tau:t})
    & =  \ f_{\theta}(\mathbf{o}_{t-\tau:t}), \\
    \mathbf{o}_{t-\tau:t} & = \ \psi(\mathbf{y}_{t-\tau:t}),
    \label{eq:vp}
\end{align}
\label{eq:image_model}
\end{subequations}
where $f_{\theta}(\cdot)$ is a DNN model, and $\psi(\cdot)$ is the conversion function to convert the lidar measurements to the binary OGMs in the robot's local frame (\ie using ray tracing). 
From this image-based model~\eqref{eq:image_model}, we can see that image-based methods explicitly ignore the kinematics and dynamics of the robot and surrounding objects (\ie they assume motion can be implicitly captured by powerful network architectures and enough good data), and fail to provide a range of possible and reliable OGM predictions (\ie they assume a deterministic future).

\subsection{SCOPE++}
\label{subsec:scope++}
Based on the limitations of image-based methods above, we argue that: 1) the future state of the environment explicitly depends on the motion of the robot itself, the motion of dynamic objects, and the state of static objects within the environment; 2) the future state of the environment is stochastic and unknown, and a range of possible future states helps provide robust predictions. 
With these two assumptions, we fully and explicitly exploit the kinematic and dynamic information of these three different types of objects, utilize the VAE-based network to provide stochastic predictions, and finally propose our novel stochastic OGM predictors SCOPE series (shown in \cref{fig:scope}) to predict the future state of the environment. 
We first create a variant called SCOPE++ that explicitly separates out static objects.
Since SCOPE++ is a superset of SCOPE, the rest of this section will focus on SCOPE++.
The prediction model~\eqref{eq:pf} of our SCOPE++, outlined in \cref{fig:scope}, can be rewritten as
\begin{subequations}
\begin{align}
    \mathbf{o}_{t+1} \sim p_{\theta}(\mathbf{o}_{t+1} \mid \mathbf{d}_{t-\tau:t})
    = & \ p_{\theta}(\mathbf{o}_{t+1} \mid \mathbf{\hat o}_{t+1}, \mathbf{m}),
    \label{eq:vae} \\
    \mathbf{\hat o}_{t+1} = \ \kappa(\mathbf{o}_{t-\tau:t}), & \  
    \mathbf{o}_{t-\tau:t} = \ \psi({\mathbf{y}^R_{t-\tau:t}}),
    \label{eq:dynamic} \\
    \mathbf{m} = & \ g({\mathbf{y}^R_{t-\tau:t}}),
    \label{eq:static} \\
    \mathbf{y}^R_{t-\tau:t} = & \ \Lambda(\mathbf{d}_{t-\tau:t}).
    \label{eq:robot} 
\end{align}
\label{eq:scope}
\end{subequations}
In this framework, \eqref{eq:vae} is the VAE predictor module from which we sample future maps $\mathbf{o}_{t+1}$.
\eqref{eq:dynamic} represent the dynamic object module, where $\kappa(\cdot)$ processes the time series data for dynamic objects, $\psi(\cdot)$ is the conversion function like \eqref{eq:vp}, and $R$ denotes the local coordinate frame of the robot at predicted time $t+n$.
\eqref{eq:static} is the static objects module, where $g(\cdot)$ is the occupancy grid mapping function for static objects. 
Finally, \eqref{eq:robot} is the robot motion module, where $\Lambda(\cdot)$ is the transformation function to compensate for robot motion.

Note that~\eqref{eq:scope} only predicts future states at the next time step $t+1$ and is used for training.
To predict a multi-step future state at time $t+n$, we can easily utilize the autoregressive mechanism and feed the next-step prediction back $\mathbf{o}_{t+1}$ to our SCOPE/SCOPE++ network \eqref{eq:vae} for $n-1$ time steps to predict the future states at time step $t+n$.
Note that the prediction horizon $n$ could theoretically be any time step.

\subsubsection{Robot Motion Compensation \texorpdfstring{$\Lambda(\cdot)$}{^(·)}}
\label{subsec:robot_motion}
To account for robot motion, we propose a simple and effective ego-motion compensation mechanism.
We first predict the future pose of the robot at the end of our prediction horizon, time $t+n$.
To do this, we use a constant velocity motion model, as it is the most widely used motion model for tracking~\cite{baisa2020derivation} and often outperforms more state-of-the-art methods in general settings~\cite{scholler2020constant} when used on a relatively short period (\ie less than \unit[1]{s}).
Note that other more appropriate robot motion models tailored to specific robot models can be used to provide better ego-motion compensation.
This results in a predicted pose for the robot $\mathbf{x}_{t+n}$, and we use this pose as the origin of our coordinate frame $R$ (where the x-axis is forward and the y-axis is left, as is standard in mobile robotics \cite{REP103}).
We then transform all data into this frame, resulting in poses $\mathbf{x}_{t-\tau:t}^R$ and lidar scans $\mathbf{y}_{t-\tau:t}^R$.
Using this data, we can generate the sequence of OGMS $\mathbf{o}_{t-\tau:t}^R$ using the conversion function $\psi(\cdot)$ by setting each lidar scan ``hit'' to 1 and all other cells to 0 (for more details, see \cite{xie2023sogmp}).
The benefit of ego-motion compensation is that we can treat these lidar measurements from a moving lidar sensor as observations in common world-fixed frame $R$, which significantly reduces the difficulty of OGM predictions and improves accuracy.
Note this only requires the robot to have accurate odometry over the history window $\tau$ (typically on the order of \unit[1]{s}), but not accurate global localization.

\subsubsection{Dynamic Object Prediction \texorpdfstring{$\kappa(\cdot)$}{k(·)}}
\label{subsec:dynamic_objects}
Tracking and predicting dynamic objects such as pedestrians is the hardest part of environmental prediction in complex dynamic scenes.
It requires some techniques to process a set of time series data to capture the motion information. 
While the traditional particle-based methods~\cite{nuss2018random} require explicitly tracking objects and treat each grid cell as an independent state, recent learning-based methods \cite{itkina2019dynamic, toyungyernsub2021double, schreiber2019long, schreiber2020motion, lange2021attention, schreiber2021dynamic, dequaire2018deep, song20192d, mohajerin2019multi} prefer to use RNNs to directly process the observed time series OGMs. 
Based on these trends, we choose the most popular ConvLSTM unit to process the spatiotemporal OGM sequences ${\mathbf{o}_{t-\tau:t}}$.
However, while other works~\cite{guen2020disentangling, toyungyernsub2021double} explicitly decouple the dynamic and static/unknown objects and use different networks to process them separately, we argue that the motion of dynamic objects is related to their surroundings, and that explicit disentangling may lose some useful contextual information.
For example, pedestrians walking through a narrow corridor are less likely to collide with or pass through surrounding walls. 
To exploit the useful contextual information between dynamic objects and their surrounding, we directly feed the observed OGMs $\mathbf{o}_{t-\tau:t}^R$ into a ConvLSTM unit $\kappa(\cdot)$ and implicitly predict the future state ${\mathbf{\hat{o}}_{t+1}}$ of dynamic (and static) objects.\footnote{Note that since the VAE is designed to provide uncertainty estimates of future states (while the static object module is designed to provide static information), our supervised learning mechanism will implicitly regularize the output of the ConvLSTM unit to contain the future states of dynamic objects. }

\begin{table*}[t]
    \small\sf\centering
    \caption{Running performance Percentage of SCOPE++ predictor on the embedded device}
    \scalebox{0.85}{
        \begin{tabular}{c c c c c}
            \toprule
            \textbf{Metrics} 
            & \textbf{Ego-Motion Compensation Module} 
            & \textbf{Static Object Module} 
            & \textbf{Dynamic Object Module} & 
            \textbf{VAE Module}\\
            
            \midrule
            Runtime & 9.04\% & \textbf{53.41\%} & 20.93\% & 16.62\% \\
            
            \midrule
            Memory Usage  & 0.32\% & 1.28\% & 48.38\% & \textbf{50.02\%} \\
           \bottomrule
        \end{tabular}
    }
    \label{tab:running_performance}
\end{table*}

\subsubsection{Static Object Segmentation \texorpdfstring{$g(\cdot)$}{g(·)}}
\label{subsec:static_objects}
While predicting dynamic objects plays a key role in environmental prediction, paying extra attention to static objects is also important to improve prediction accuracy. 
The main reason is that the area occupied by static objects is much larger than that of dynamic objects, as shown in \cref{fig:scope}, where static objects such as walls are much larger than dynamic objects such as pedestrians, which are sparse and scattered point clusters.
Another reason is that static objects maintain their shape and position over time, contributing to the scene geometry and giving a global view of the surroundings.
To account for static objects, we utilize a local static environment map $\mathbf{m}$ as a prediction for future static objects, which is a key contribution of our work. 
We generate this local environment map $\mathbf{m}$ using a GPU-accelerated implementation\footnote{\url{https://github.com/TempleRAIL/occupancy_grid_mapping_torch}} of the standard inverse sensor model~\cite{thrun2003learning}  for $g(\cdot)$ that parallelizes independent cell state update operations by taking advantage of the additive update rule (for more details, see \cite{xie2023sogmp}).
This Bayesian approach generates a robust estimate of the local map, where dynamic objects are treated as noise data and removed over time. 

\subsubsection{VAE Predictor}
\label{subsec:vae_predictor}
The dynamic and static prediction functions from \eqref{eq:dynamic} and \eqref{eq:static} yield one estimate of the future environment.
However, it is unlikely that this exact map will be correct.
To account for this, we wish to generate a range of possible future environment states that capture the uncertainty of the environment.
We do this using a variational autoencoder (VAE), a generative machine learning technique that creates samples that are ``similar'' to the predicted OGM $\mathbf{\hat o}_{t+1}$.

To represent the stochasticity of environment states, we assume that environment states $\mathbf{o}_{t-\tau:t+n}$ are generated by some unobserved, random, latent variable $\mathbf{z}$ that follow a prior distribution $p_{\theta}(\mathbf{z})$.
Our VAE model~\eqref{eq:vae} can be rewritten as
\begin{equation}
    p_{\theta}(\mathbf{o}_{t+1} \mid \mathbf{\hat o}_{t+1}, \mathbf{m})
    = \int p_{\theta}(\mathbf{z}) p_{\theta}(\mathbf{o}_{t+1} \mid  \mathbf{z}, \mathbf{\hat o}_{t+1}, \mathbf{m}) dz. 
    \label{eq:marginal_likelihood} 
\end{equation}
Since we are unable to directly optimize this marginal likelihood and obtain optimal parameters $\theta$, we use a VAE network to parameterize our prediction model $p_{\theta}(\mathbf{o}_{t+1} \mid \mathbf{\hat o}_{t+1}, \mathbf{m})$, outlined in \cref{fig:scope}, where the inference network (encoder) parameterized by $\phi$ refers to the variational approximation $q_{\phi}(\mathbf{z} \mid \mathbf{\hat o}_{t+1}, \mathbf{m})$, the generative network (decoder) parameterized by $\theta$ refers to the likelihood $ p_{\theta}(\mathbf{o}_{t+1} \mid \mathbf{z}, \mathbf{\hat o}_{t+1}, \mathbf{m})$, and the standard Gaussian distribution $\mathcal{N} (0, 1)$ refers to the prior $p_{\theta}(\mathbf{z})$.
Then, from work~\cite{kingma2013auto}, we can simply maximize the evidence lower bound (ELBO) loss $\mathcal{L} (\theta, \phi; \mathbf{o}_{t+1})$ to optimize this marginal likelihood and get the optimal $\theta$:
\begin{equation}
    \begin{split}
        \mathcal{L} (\theta, \phi; \mathbf{o}_{t+1}) = & 
        \mathbb{E}_{q_{\phi}(\mathbf{z} \mid \mathbf{\hat o}_{t+1}, \mathbf{m})} \left[\log{ p_{\theta}(\mathbf{o}_{t+1} \mid  \mathbf{z}, \mathbf{\hat o}_{t+1}, \mathbf{m})} \right] \\
        & -
        KL \big(
        q_{\phi}(\mathbf{z} \mid \mathbf{\hat o}_{t+1}, \mathbf{m})  \,\|\,
        p_{\theta}(\mathbf{z}) 
        \big).
    \end{split}
    \label{eq:elbo} 
\end{equation}
The first term on the right-hand side (RHS) is the expected generative error, describing how well the future environment states can be generated from the latent variable $\mathbf{z}$.
The second RHS term is the Kullback–Leibler (KL) divergence, describing how close the variational approximation is to the prior.

We use mini-batching, Monte-Carlo estimation, and reparameterization tricks to calculate the gradients of the ELBO~\eqref{eq:elbo}~\cite{kingma2013auto}, and obtain the optimized model parameters $\phi$ and $\theta$. 
Finally, our VAE can integrate the predicted features $\{\mathbf{\hat o}_{t+1}, \mathbf{m} \}$ of dynamic and static objects, and output a probabilistic estimate of future OGM states with uncertainty awareness. 

\section{Software Optimization}
\label{sec:software_acceleration}
We demonstrated in our previous work~\cite{xie2023sogmp} (results also in \cref{sec:ogm_results}) that SCOPE and SCOPE++ can accurately predict OGMs.
However, the use of multiple neural networks is resource-intensive.
For example, the VAE used to generate samples to provide uncertainty estimates is so memory-intensive that we can only generate 8 samples using an NVIDIA Jetson AVG Xavier embedded device~\cite{xie2023sogmp}.\footnote{Generating 8 samples is a trade-off in running speed and memory consumption when running with other navigation-related packages such as \texttt{amcl}, \texttt{move\_base}, and other sensor driver ROS packages.}
To make our proposed SCOPE/SCOPE++ work efficiently in resource-constrained robots, we decide to optimize them on the software side.

\subsection{Resource Utilization}
\Cref{tab:running_performance} details the running performance percentage of each module in the SCOPE++ predictor (generating 8 samples) in terms of runtime and memory usage on our embedded computer.
We can easily see that the runtime is mainly limited by the static and dynamic object modules (with the ConvLSTM unit), while the memory usage is mainly limited by the dynamic object (with ConvLSTM unit) and VAE modules. 
This analysis explains why our proposed SCOPE predictor without the static object module is much faster than our SCOPE++ predictor. 

Since the key ConvLSTM unit in the dynamic object module is already an accelerated RNN unit (compared to the Vanilla RNN unit), it is difficult to further optimize this module.
Therefore, we choose to perform software optimization on the static object module and VAE module to increase running speed and reduce memory consumption while maintaining the accurate prediction and stochasticity of our SCOPE/SCOPE++.

According to \Cref{tab:running_performance} and our network architecture design, a straightforward method is to remove the static object module (\ie use SCOPE instead of SCOPE++), especially since the two variants achieve comparable prediction accuracy.
Next, we will look at replacing the VAE network by utilizing knowledge distillation techniques to compress our SCOPE network to maintain its prediction performance, and then quantify the prediction uncertainty of the SCOPE to provide uncertainty estimates to maintain its stochastic nature.

\subsection{Knowledge Distillation}
\label{subsec:knowledge_distillation}
Compared with other deep neural network compression techniques (\eg pruning, quantization), knowledge distillation techniques can provide better generalization capabilities and allow the selection of different network architectures~\cite{stanton2021does}.
To design a simple and efficient student network for knowledge distillation, we remove the static object module (following our fast SCOPE network architecture) and replace the complete VAE network with a single convolutional layer, which can be seen from the SO-SCOPE part of \cref{fig:scope}.
Therefore, our SO-SCOPE network is very simple and efficient, consisting of only one ConvLSTM layer and one convolutional layer.

We use the SCOPE instead of the SCOPE++ as our teacher network for knowledge distillation, shown in \cref{fig:scope}. 
There are two reasons why we choose the faster and less accurate SCOPE network rather than the complex and more accurate SCOPE++ network as the teacher network.
First, as Cho et al.~\cite{cho2019efficacy} suggested, larger or more accurate models generally do not make better teachers in the knowledge distillation, especially when the student's ability is too low to successfully imitate the teacher.
Second, Beyer et al.~\cite{beyer2022knowledge} suggested that keeping ``consistent'' teaching can improve knowledge distillation. Our SO-SCOPE student network follows the SCOPE network design, which is almost the same as the SCOPE network except that a single convolutional layer is used instead of the VAE network (\ie the input view of the convolutional layer is consistent with the VAE network).
Therefore, this structural similarity and consistency improves knowledge transfer from the teacher network to the student network.
This will accelerate the running speed and also reduce the memory consumption.

To train our SO-SCOPE student network, we directly use the output of our trained SCOPE teacher model in the OGM-Turtlebot2 training dataset (see \cref{subsubsec:datasets}) as ``soft'' labels instead of using ground truth labels, which follow the key idea of knowledge distillation. 
To speed up training, we use 1 random sample output from SCOPE as a ``soft'' label in each training iteration.
Note that since our SCOPE is stochastic and generates different samples even with the same input, this stochastic sample output expands the original training dataset (\ie data augmentation) and further improves the generalization ability of the SO-SCOPE.
We use the binary cross entropy (BCE) loss function to train our SO-SCOPE student network and get our final SO-SCOPE predictor to provide deterministic OGM predictions. 
After completing the knowledge distillation training, we directly use the output of the SO-SCOPE predictor to provide OGM prediction information.
However, this distilled network is now deterministic, \ie each input will generate a single output.
Our next step will address this limitation.

\begin{figure*}[t]
    \centering
    \subfloat[$T = 1$, $\hat{c}_{t+1} \in {\left[0, \frac{1}{15}\right]}$]{
            \centering
            \includegraphics[width=0.18\textwidth]{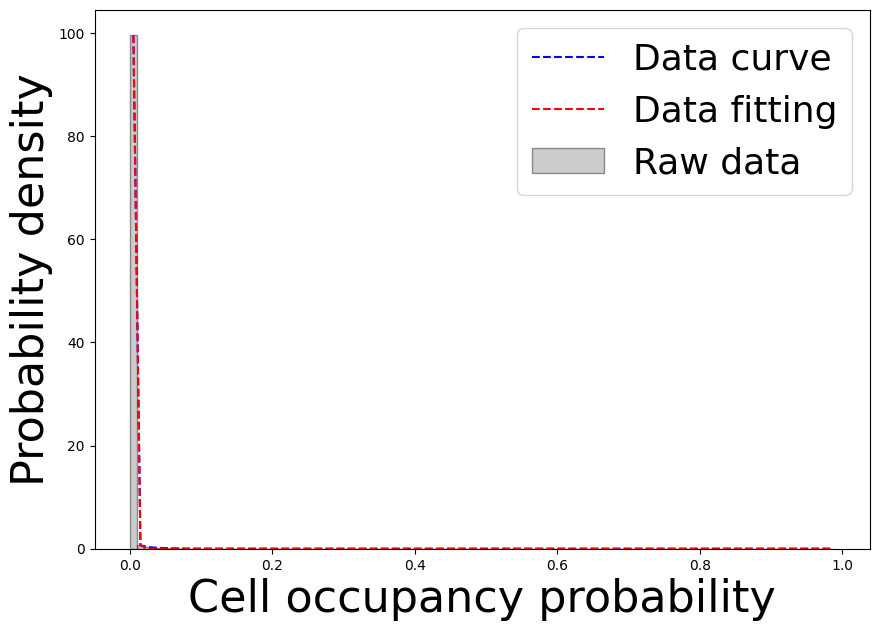}
            \label{fig:1th_timestep_bin0}
    }%
    \subfloat[$T = 1$, $\hat{c}_{t+1} \in {\left[\frac{2}{15}, \frac{3}{15}\right]}$]{
            \centering
            \includegraphics[width=0.18\textwidth]{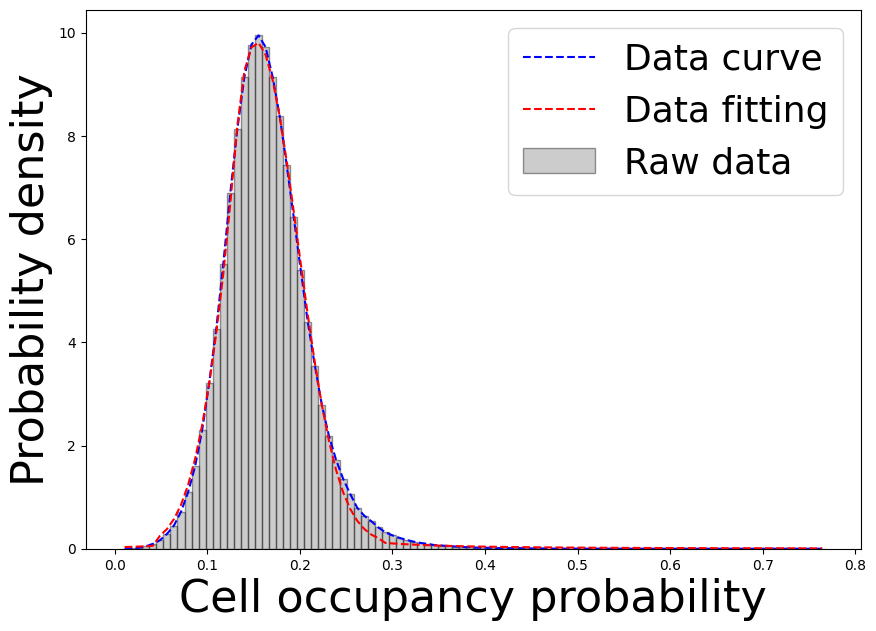}
            \label{fig:1th_timestep_bin2}
    }%
    \subfloat[$T = 1$, $\hat{c}_{t+1} \in {\left[\frac{5}{15}, \frac{6}{15}\right]}$]{
            \centering
            \includegraphics[width=0.18\textwidth]{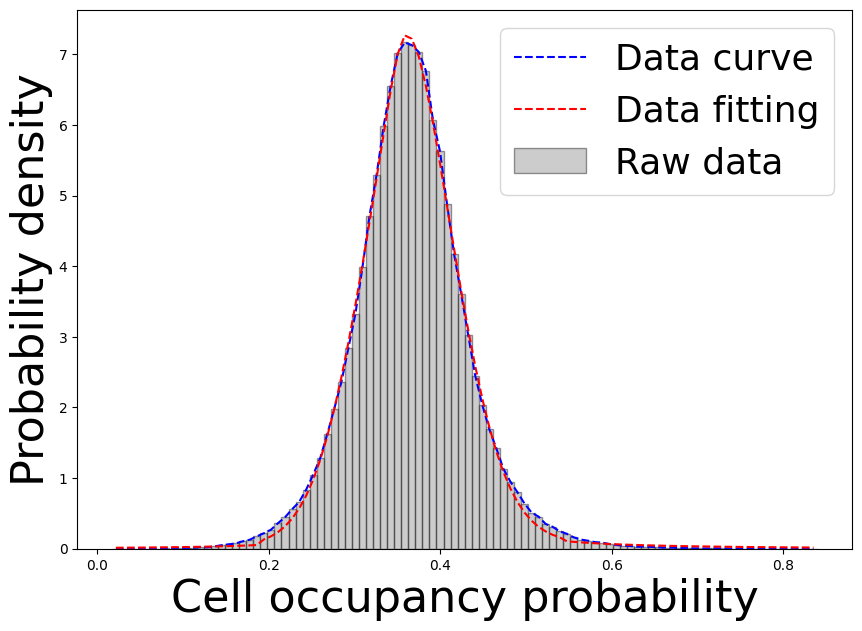}
            \label{fig:1th_timestep_bi5}
    }%
    \subfloat[$T = 1$, $\hat{c}_{t+1} \in {\left[\frac{10}{15}, \frac{11}{15}\right]}$]{
            \centering
            \includegraphics[width=0.18\textwidth]{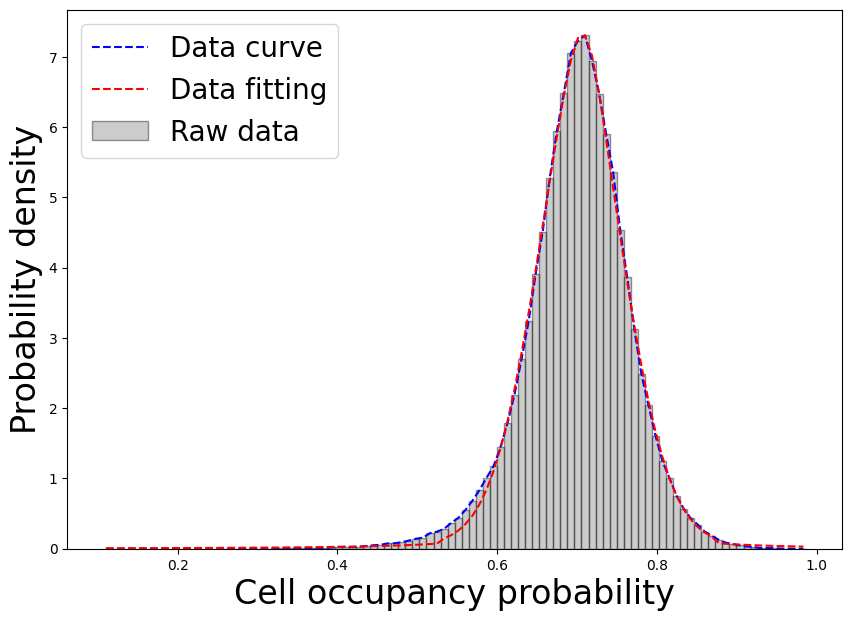}
            \label{fig:1th_timestep_bin10}
    }%
    \subfloat[$T = 1$, $\hat{c}_{t+1} \in {\left[\frac{14}{15},1\right]}$]{
            \centering
            \includegraphics[width=0.18\textwidth]{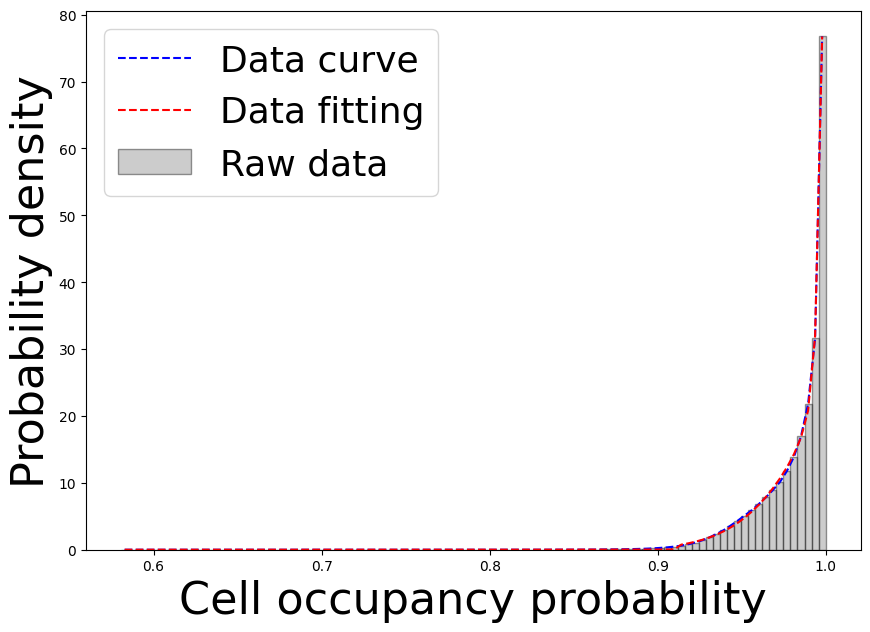}
            \label{fig:1th_timestep_bin14}
    }%
    \vspace{0.1pt}
    \subfloat[$T = 5$, $\hat{c}_{t+1} \in {\left[0, \frac{1}{15}\right]}$]{
            \centering
            \includegraphics[width=0.18\textwidth]{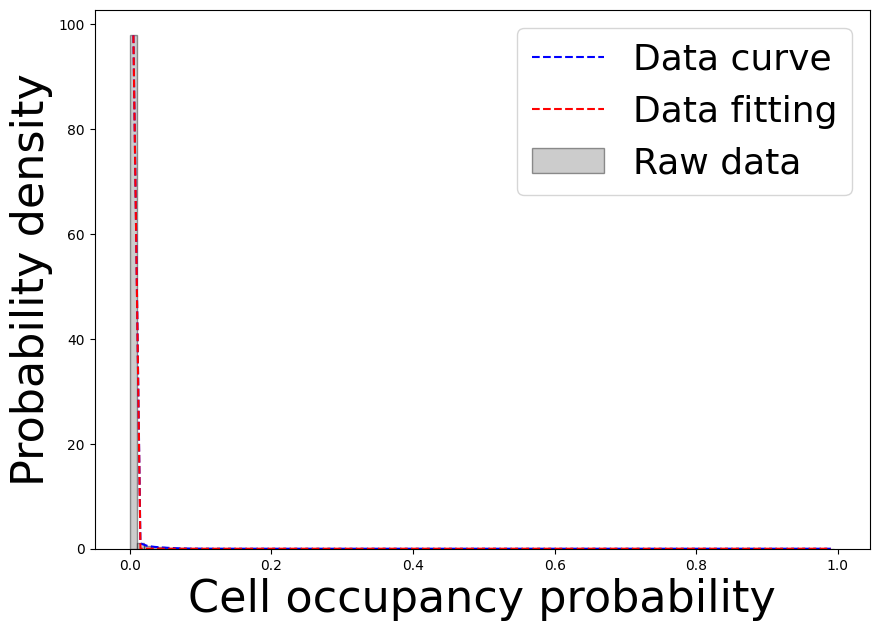}
            \label{fig:5th_timestep_bin0}
    }%
    \subfloat[$T = 5$, $\hat{c}_{t+1} \in {\left[\frac{2}{15}, \frac{3}{15}\right]}$]{
            \centering
            \includegraphics[width=0.18\textwidth]{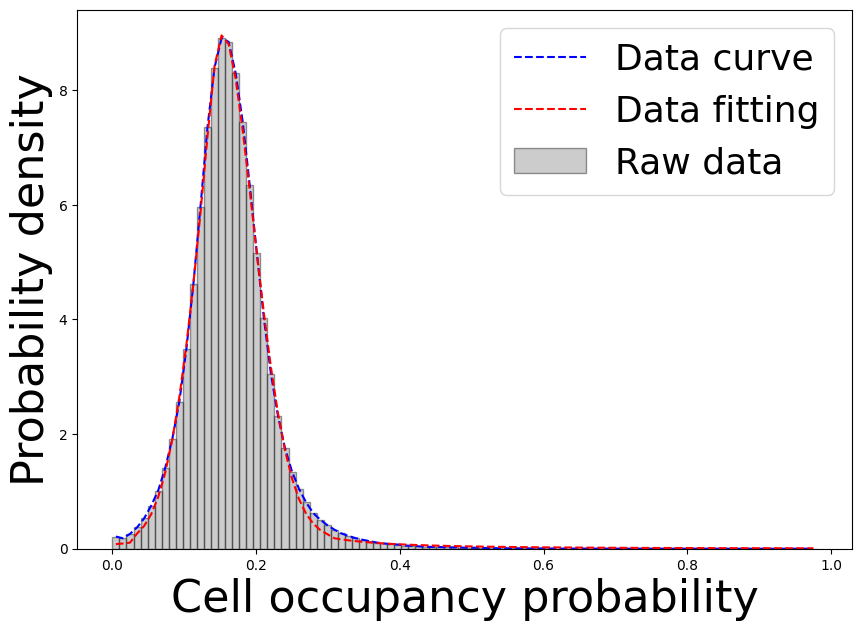}
            \label{fig:5th_timestep_bin2}
    }%
    \subfloat[$T = 5$, $\hat{c}_{t+1} \in {\left[\frac{5}{15}, \frac{6}{15}\right]}$]{
            \centering
            \includegraphics[width=0.18\textwidth]{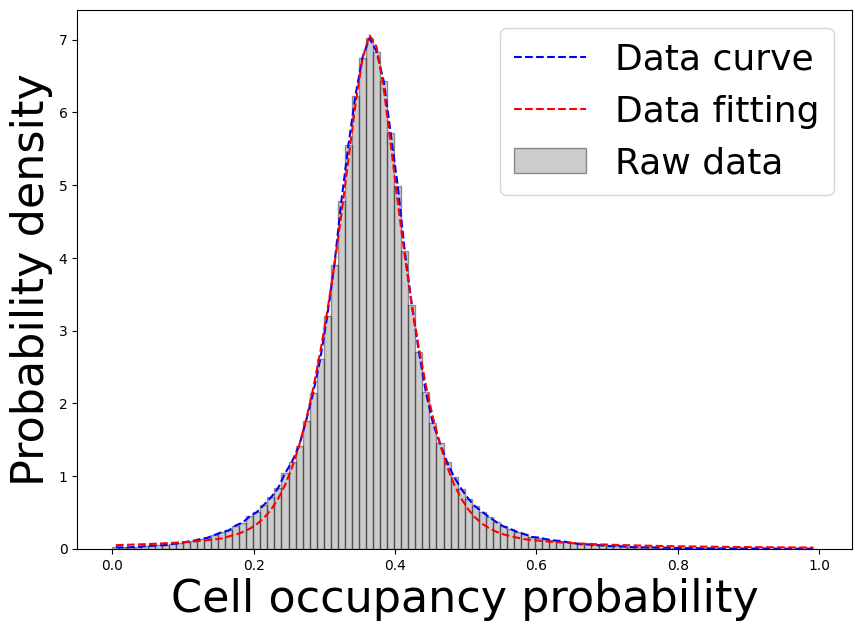}
            \label{fig:5th_timestep_bin3}
    }%
    \subfloat[$T = 5$, $\hat{c}_{t+1} \in {\left[\frac{10}{15}, \frac{11}{15}\right]}$]{
            \centering
            \includegraphics[width=0.18\textwidth]{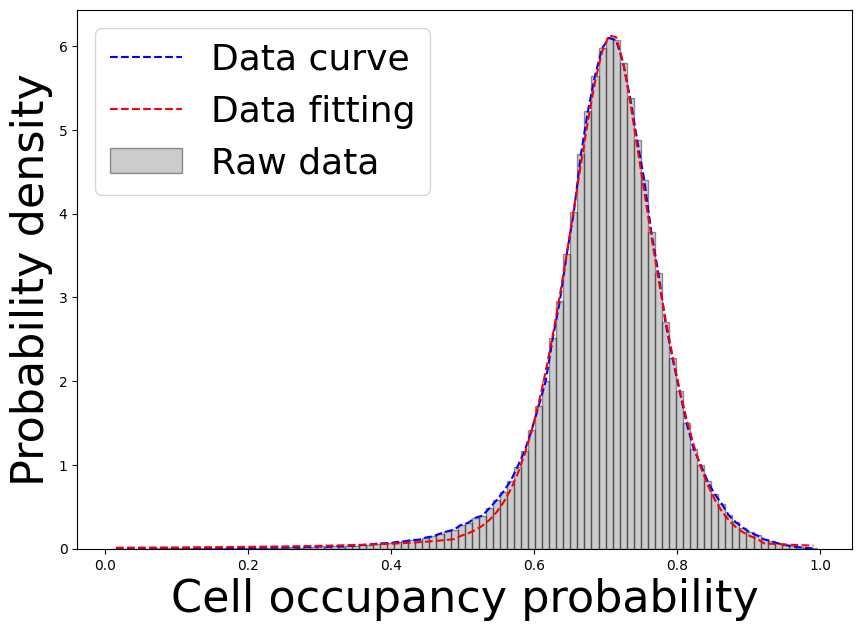}
            \label{fig:5th_timestep_bin6}
    }%
    \subfloat[$T = 5$, $\hat{c}_{t+1} \in {\left[\frac{14}{15}, 1\right]}$]{
            \centering
            \includegraphics[width=0.18\textwidth]{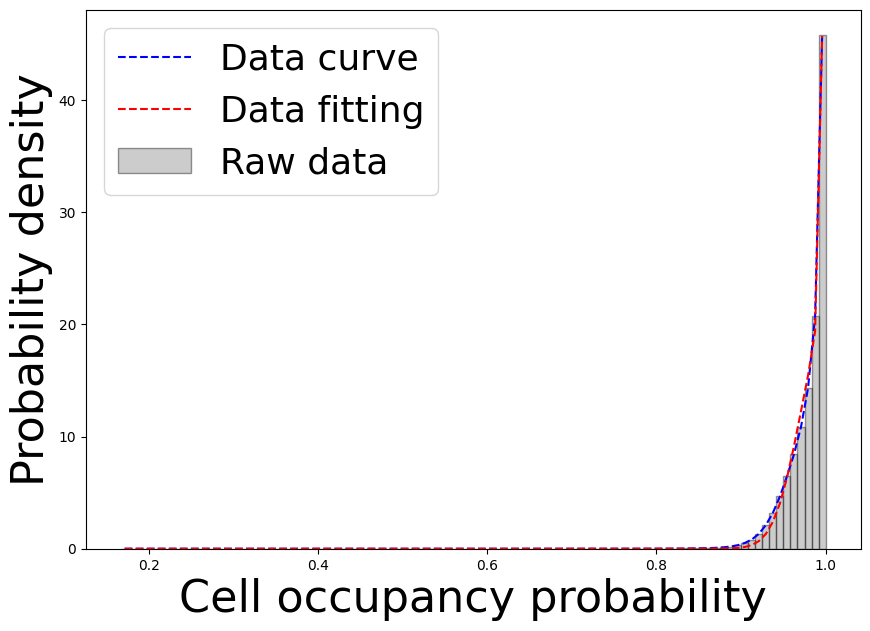}
            \label{fig:5th_timestep_bin10}
    }%
    \vspace{0.1pt}
    \subfloat[$T = 10$, $\hat{c}_{t+1} \in {\left[0, \frac{1}{15}\right]}$]{
            \centering
            \includegraphics[width=0.18\textwidth]{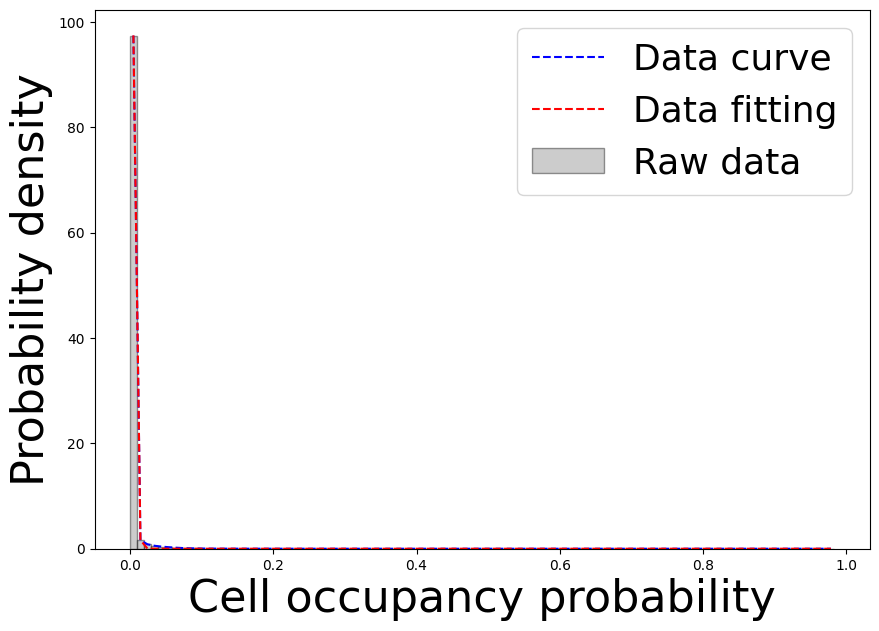}
            \label{fig:10th_timestep_bin0}
    }%
    \subfloat[$T = 10$, $\hat{c}_{t+1} \in {\left[\frac{2}{15}, \frac{3}{15}\right]}$]{
            \centering
            \includegraphics[width=0.18\textwidth]{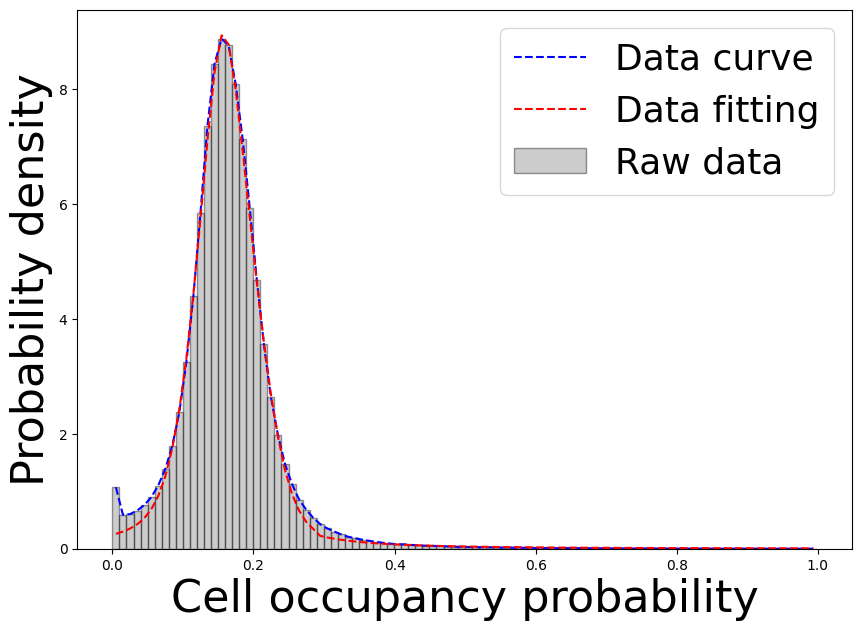}
            \label{fig:10th_timestep_bin2}
    }%
    \subfloat[$T = 10$, $\hat{c}_{t+1} \in {\left[\frac{5}{15}, \frac{6}{15}\right]}$]{
            \centering
            \includegraphics[width=0.18\textwidth]{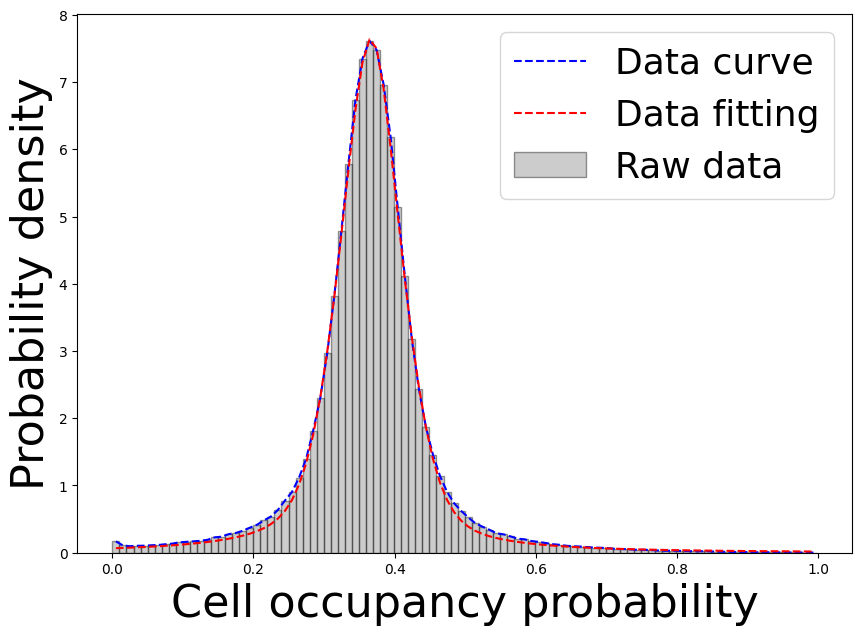}
            \label{fig:10th_timestep_bin3}
    }%
    \subfloat[$T = 1$, $\hat{c}_{t+1} \in {\left[\frac{10}{15}, \frac{11}{15}\right]}$]{
            \centering
            \includegraphics[width=0.18\textwidth]{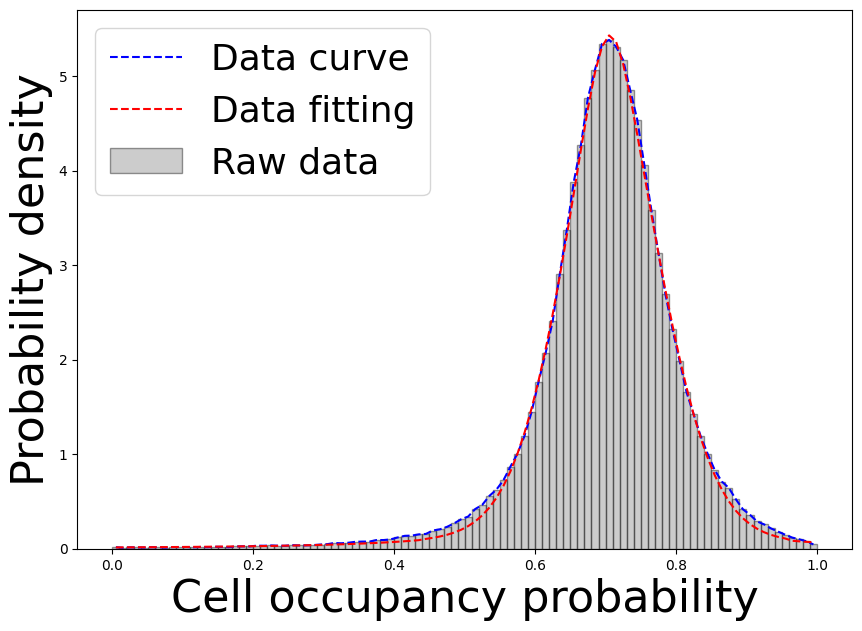}
            \label{fig:10th_timestep_bin6}
    }%
    \subfloat[$T = 1$, $\hat{c}_{t+1} \in {\left[\frac{14}{15}, 1\right]}$]{
            \centering
            \includegraphics[width=0.18\textwidth]{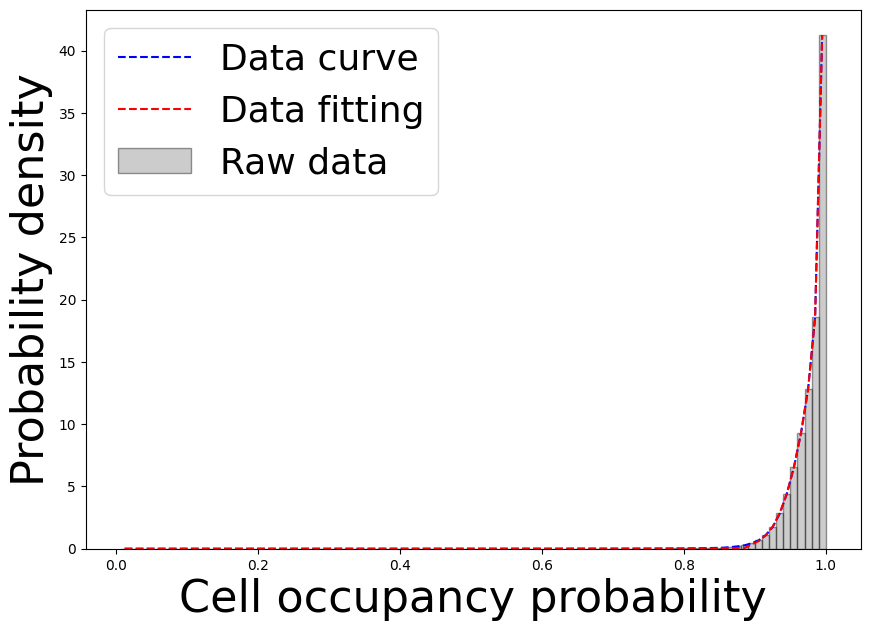}
            \label{fig:10th_timestep_bin10}
    }%
    \caption{Uncertainty quantification of the predicted OGM cell bins (each column is one occupancy bin $\tilde{c}_{t+1}$) at different prediction time steps (each row is one timestep $T$).
    Grey histograms are the raw predicted OGM cell data of our SCOPE++ predictor.
    Blue curves are the connecting curves of these raw predicted OGM cell data.
    Red curves are the fitting curves from our proposed mixture model of truncated normal and skew-Cauchy distributions in \eqref{eq:mixture_model}.
    }
    \label{fig:ogm_bin_distribution}
\end{figure*}

\subsection{Uncertainty Quantification}
\label{subsec:uncertainty_quantification}
To modify SO-SCOPE, we will characterize the statistics of the VAE from our trained SCOPE++ predictor and then combine this data with the output of the distillation network (which should generate an OGM that is similar to an output of the VAE) to draw samples with identical statistics to our original network but in a much more memory- and time-efficient manner.
Note that using SCOPE++ here is a reasonable choice because our SCOPE++ predictor provides a more accurate and comprehensive uncertainty estimate than SCOPE.
To characterize the VAE, we collected $8,000$ different input sequences to the ConvLSTM from the OGM-Turtlebot2 testing dataset (see \cref{subsubsec:datasets}) and generated 32 VAE samples for each input, giving us a total of $256,000$ output samples for each prediction horizon $T = 1, \ldots, \tau$.
Although using the OGM-Turtlebot2 testing dataset may raise the issue of data leakage, we used the testing dataset to characterize the VAE because: 1) SCOPE++ is well trained on the training dataset and, as a result, the uncertainty of SCOPE++’s output in the training dataset is very small and the uncertainty statistics in the training dataset do not reflect the true uncertainty statistics of the SCOPE++;
in contrast, the uncertainty statistics in the testing dataset reflect the true uncertainty statistics of the SCOPE++ very well; 
and 2) the key aim of proposing our software-optimized SO-SCOPE predictor is to optimize and accelerate the trained resource-intensive SCOPE/SCOPE++ predictors to allow them to run in real time on resource-constrained robots. 
It is reasonable to choose a testing dataset to characterize uncertainty statistics because we already have the predictors that need to be accelerated and we can easily use them to collect real data statistics from the application environment.

Our analysis relies on two factors.
First, in an OGM, the probability of occupancy for each cell is independent of all other cells~\cite{thrun2003learning}.
In other words, for two cells $c$ and $c'$, then $p(c\textrm{ is occupied and }c'\textrm{ is occupied}) = p(c\textrm{ is occupied}) p(c'\textrm{ is occupied})$.
Second, based on our use of a VAE, we make the assumption that the distribution of occupancy probabilities in the VAE output depends on the occupancy probability of the input. 
This is reasonable since a VAE is trained to have similar outputs near to one another in the learned latent space.
Let $\hat{c}_{t+T} = p(c\textrm{ is occupied at time }t+T)$ denote the probability of occupancy of cell $c$ predicted by the ConvLSTM layer $T$ steps into the future (\ie we use the average of the occupancy probabilities predicted by the VAE for the same cell as its estimate) and $\tilde{c}_{t+T}$ be the probability of occupancy predicted for that same cell by the VAE.
Then we assume $p(\tilde{c}_{t+T} \mid \hat{c}_{t+T}) \neq p(\tilde{c}_{t+T})$.

Based on these factors, we will examine trends in the VAE probabilities $\tilde{c}_{t+T}$ (for $T = 1, \ldots, \tau$) conditioned on the value of $\hat{c}_{t+T}$.
We use 15 bins for $\hat{c}_{t+T}$, evenly spaced from 0 to 1 (since the probability range of OGM cell is in $[0,1]$). 
With these binned OGM cell samples, we visualize the distribution of $\tilde{c}_{t+T}$ for each $\hat{c}_{t+T}$ bin, as shown in \cref{fig:ogm_bin_distribution}. 
We can see a few useful trends in the data.
First, cells that are predicted to be empty by the ConvLSTM (\ie in bin 0) remain empty for nearly all VAE samples.
This trend holds during the full prediction window, with the occupancy probability increasing only slightly as $T$ becomes larger. 
Second, the peak of the distribution for $\tilde{c}_{t+T}$ remains very close to the value of $\hat{c}_{t+1}$.
This implies that the VAE generates a consistent distribution around the ConvLSTM output.
Third, heavier tails in the distributions as $T$ increases.
This makes sense as a longer horizon into the future leads to further uncertainty and further deviation from the current time.
Fourth, the distributions skew towards 0 as the prediction horizon $T$ increases.
This makes sense as cells are more likely to become empty as people move about in the scene.
Fifth, for situations where $\hat{c}_{t+1}$ is low, we see a peak in the distribution of $\tilde{c}_{t+T}$ at 0.
This is caused by uncertain objects moving out of a cell, leaving it empty.

Our goal is to fit a model to the resulting histograms.
Our model must be able to capture the heavy tails, skew, and the peak near $\tilde{c} = 0$.
Based on this, we use a mixture model of a truncated normal and a skew-Cauchy distribution to describe the distribution of $\tilde{c}_{t+T}$ values, which can be expressed as:
\begin{equation}
    \begin{split}
        f_{\mathbf{\xi}}(\tilde{c}) = & 
         w  \frac{\sqrt{\frac{2}{\pi \sigma_{\rm tn}^2}} \exp\left( -\frac{1}{2} \left(\frac{\tilde{c}-\mu_{\rm tn}}{\sigma_{\rm tn}} \right)^2\right)}{\textrm{erf}\left(\frac{ b-\mu_{\rm tn}}{\sqrt{2} \sigma_{\rm tn}}\right) - \textrm{erf}\left(\frac{a -\mu_{\rm tn}}{\sqrt{2} \sigma_{\rm tn}}\right)} + \\
        & (1-w) \frac{1}{\sigma_{sc} \pi \left[ 1 + \left( \frac{|\tilde{c}-\mu_{sc}|}{\sigma_{sc} \left( 1 + \lambda \, \textrm{sign}(\tilde{c}-\mu_{sc}) \right)} \right)^2 \right]},
    \end{split}
    \label{eq:mixture_model}
\end{equation}
where the first term on RHS is the truncated normal distribution (with mean $\mu_{\rm tn}$, standard deviation $\sigma_{\rm tn}$, lower bound $a$, and upper bound $b$) and the second RHS term is the skew-Cauchy distribution (with location parameter $\mu_{\rm sc}$, scale parameter $\sigma_{\rm sc}$, and skewness $\lambda$), and $w \in [0,1]$ is the mixture weight.

Let $\mathbf{\xi} = [w, a, b,\mu_{\rm tn}, \sigma_{\rm tn}, \lambda, \mu_{\rm sc}, \sigma_{\rm sc}]$ be the vector of mixture model parameters. 
We use a least square optimization algorithm~\cite{beale1960confidence} to fit $\mathbf{\xi}$ to our predicted OGM cell bin data.
\Cref{fig:ogm_bin_distribution} shows the results of this fit.
We see that the model achieves strong performance across a range of bins and timesteps, indicating that our methods work well.

The end result of our analysis above is a set of 150 $\mathbf{\xi}$ vectors, one for each of the 15 $\hat{c}$ bins and each of the 10 time steps $T$.
For each set of mixture model parameters, we can generate different types of distribution statistics (\eg mean, median, mode, entropy, and random variate) and store these values in a lookup table.
We call this lookup table the prediction uncertainty statistics lookup table.

Now that we have accurate models for the statistics of how each cell in the map will change over time, we can integrate that information into the SO-SCOPE model.
To do this, we use the output of the distillation layer to get $\hat{c}_{t+T}$ for each cell in the OGM $\mathbf{o}_{t+T}$, as shown in the SO-SCOPE part of \cref{fig:scope}.
We can then use the uncertainty statistics lookup table to get statistics of that cell, or we can use the cumulative distribution function of \eqref{eq:mixture_model} to draw samples $\tilde{c}_{t+T}$.
This allows our SO-SCOPE model to account for uncertainty in the predicted future using knowledge distilled from the SCOPE and SCOPE++ models to achieve comparable accuracy with significantly smaller computational effort and memory consumption.

\section{OGM Prediction Experiments and Results}
\label{sec:ogm_results}
To demonstrate the prediction performance of our proposed approaches, we first test our algorithms on a simulated dataset and two public real-world sub-datasets from the socially compliant navigation dataset (SCAND)~\cite{karnan2022socially}. 
We comprehensively evaluate the inference speed and memory usage of our proposed predictors on a resource-constrained embedded computing device to show that our software-optimized predictor can achieve surprising performance improvements.
We also characterize the uncertainty of our proposed predictors across different sample sizes and numbers of objects to demonstrate the diversity and consistency of their uncertainty estimates.

\subsection{Datasets}
\label{subsubsec:datasets}
We used three publicly available OGM datasets to evaluate our proposed prediction algorithms and baselines, one self-collected from the Gazebo simulator (\ie OGM-Turtlebot2)~\cite{sogmp_dataset} and two extracted from the real-world dataset SCAND~\cite{karnan2022socially} (\ie OGM-Jackal and OGM-Spot). 

\subsubsection{OGM-Turtlebot2 Dataset}
\label{subsubsec:ogm_turtlebot}
To train our predictors and baselines, we collected the OGM-Turtlebot2 dataset using the 3D robot-pedestrian interaction Gazebo simulator from our previous work~\cite{xie2021towards, xie2023drlvo}. 
We collected the robot states $\{\mathbf{x},\mathbf{u}\}$ and raw lidar measurements $\mathbf{y}$ at a sampling rate of \unit[10]{Hz}.
We collected a total of 94,891 $\mathbf{d} = \{\mathbf{x},\mathbf{u},\mathbf{y}\}$ data tuples, dividing this into three separate subsets for training (67,000 tuples), validation during training (10,891 tuples), and final testing (17,000 tuples).
More dataset details can be found in~\cite{xie2023sogmp}.

\subsubsection{OGM-Jackal and OGM-Spot Datasets}
\label{subsubsec:ogm_jackal_spot}
The OGM-Jackal and OGM-Spot datasets were processed and created from the original SCAND dataset~\cite{karnan2022socially}, which was collected by humans manually controlling the Jackal robot and the Spot robot around the indoor/outdoor flat environments at UT Austin.\footnote{Force control for complex dynamic motions like the Spot robot is handled by the low-level controller, while the human motion control commands collected in the dataset are high-level 2D velocity control commands.}
More dataset processing details can be found in~\cite{xie2023sogmp}.

\subsection{OGM Prediction}
\label{sec:prediction_results}
We compare our proposed SCOPE, SCOPE++, and SO-SCOPE algorithms with six deep learning-based baselines: ConvLSTM~\cite{shi2015convolutional}, DeepTracking~\cite{ondruska2016deep}, PhyDNet~\cite{guen2020disentangling}, SAAConvLSTM~\cite{lange2021attention}, TAAConvLSTM~\cite{lange2021attention}, and LOPR~\cite{lange2022lopr}, and six ablation baselines: SCOPE-NEMC (SCOPE with No Ego-Motion Compensation module), SO-SCOPE-GTEMC (SO-SCOPE with Ground Truth of Ego-Motion Compensation module), SO-SCOPE-MEAN (use the mean value from the prediction uncertainty statistics lookup table as the OGM prediction), SO-SCOPE-MEDIAN (use the median from the lookup table), SO-SCOPE-MODE (use the mode from the lookup table), and SO-SCOPE-SAMPLE (draw a random sample from distribution)~\footnote{Note that we ignore two ablation baselines of training the same SO-SCOPE network using dataset labels and using the SCOPE++ model as the ``teacher'' network, as they are both used to support the knowledge distillation theory and validated by previous researchers~\cite{stanton2021does, cho2019efficacy, beyer2022knowledge}.}.
Note that SCOPE-NEMC is used to demonstrate the advantages of our proposed ego-motion compensation module and the network structure compared to other image-based baselines without ego-motion compensation.
SO-SCOPE-GTEMC is used to demonstrate the upper-performance limit of the SO-SCOPE predictor with accurate ego-motion compensation.
All these networks were implemented using PyTorch framework~\cite{paszke2019pytorch} and trained using \textit{only} our self-collected OGM-Turtlebot2 dataset.

\subsubsection{Evaluation Metrics}
\label{subsubsec:evaluation_metrics}
We use three metrics: 1) weighted mean square error (WMSE)~\cite{ponomarenko2010weighted} to evaluate absolute error between the ground truth and predicted maps, 2) structural similarity index measure (SSIM)~\cite{wang2004image} to measure the similarity between the two maps, and 3) optimal subpattern assignment metric (OSPA)~\cite{Schuhmacher2008Metric} to measure the difference in the number and locations of distinct objects within the maps.  
Note that OGM prediction works~\cite{itkina2019dynamic, schreiber2019long, schreiber2020motion, lange2021attention, lange2022lopr, toyungyernsub2021double, schreiber2021dynamic, dequaire2018deep, song20192d, mohajerin2019multi} only use the computer vision metrics (\eg MSE, F1 Score, and SSIM) to evaluate the quality of predicted OGMs.
Since a blurred prediction with uncertainty may perform better in some cases, we are the first to evaluate the predicted OGMs from the perspective of multi-target tracking (\ie OSPA distance), which we believe gives a more physically meaningful evaluation than image quality as it accounts for the number of and locations of objects in the scene.
More details can be found in~\cite{xie2023sogmp}.

\subsubsection{Qualitative Results}
\label{subsubsec:qualitative_results}
\begin{figure*}[t]
    \centering
    \includegraphics[width=0.78\textwidth]{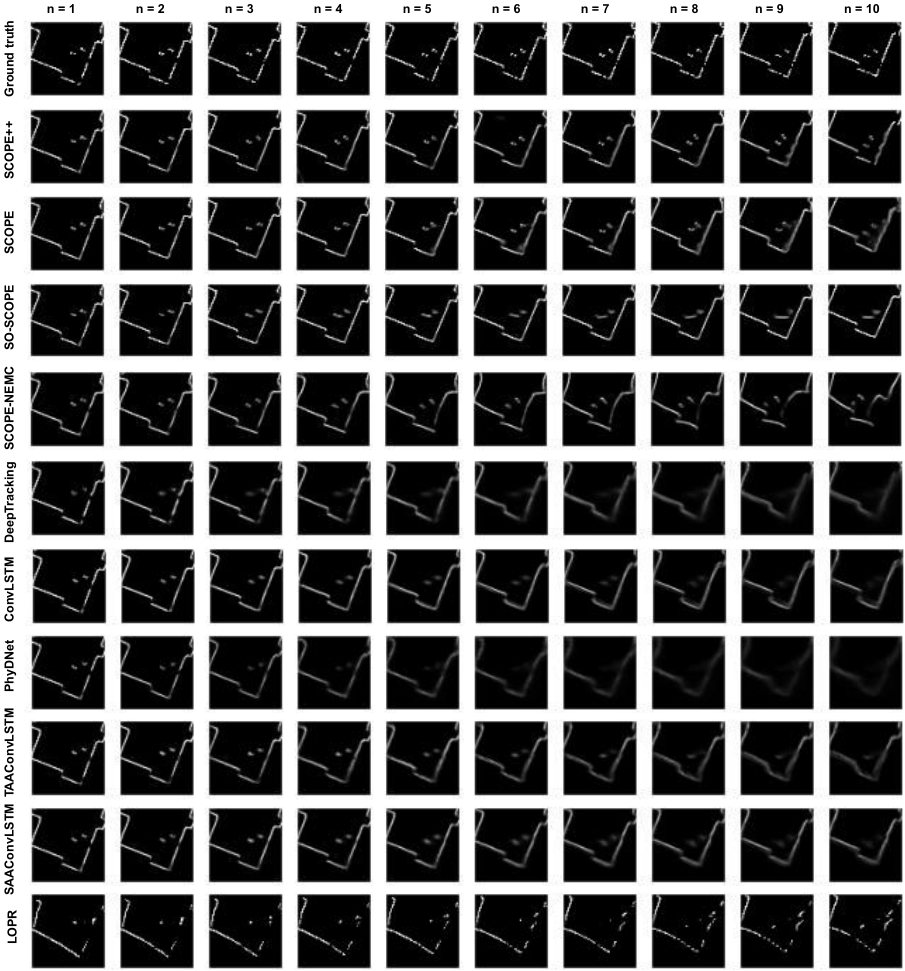}
    \caption{Prediction showcase of ten OGM predictors tested on the OGM-Turtlebot2 dataset. 
    The black and white areas are free and occupied space respectively.
    }
    \label{fig:showcase}
\end{figure*}

\Cref{fig:showcase} and the accompanying multimedia illustrate the future OGM predictions generated by our proposed predictors and the baselines.
We observe three interesting phenomena.
First, the image-based baselines, especially the PhyDNet, generate blurry future predictions after 5-time steps, with only blurred shapes of static objects (\ie walls) and missing dynamic objects (\ie pedestrians).
We believe this is because these six baselines are deterministic models that use less expressive network architectures, only treat time series OGMs as images/video, and cannot capture and utilize the kinematics and dynamics of the robot itself, dynamic objects, and static objects.
Second, the SCOPE++ with a local environment map has a sharper and more accurate surrounding scene geometry (\ie right walls) than the SCOPE without it. 
This difference indicates that the local environment map for static objects is beneficial and plays a key role in predicting surrounding scene geometry.
Third, our proposed software-optimized SO-SCOPE can achieve clear and sharp OGM predictions similar to its ``teacher'' SCOPE, which demonstrates the effectiveness of applying knowledge distillation techniques to optimize our SCOPE/SCOPE++.

\subsubsection{Quantitative Results}
\label{subsubsec:quantitative_results}
While achieving sharper predictions is encouraging, we also need quantitative measures. 
We test 14 OGM predictors on the OGM-Turtlebot2 test dataset, OGM-Jackal dataset, and OGM-Spot dataset and provide a comprehensive benchmark by evaluating absolute error, structure similarity, and tracking accuracy.
Recall that we only use the OGM-Turtlebot2 during training, so the other two datasets are entirely new and were generated by robots with different speeds and kinematics (\ie Spot ambulates).

\begin{figure*}[t]
    \centering
    \subfloat[WMSE: OGM-Turtlebot2]{
            \centering
            \includegraphics[width=0.3\textwidth]{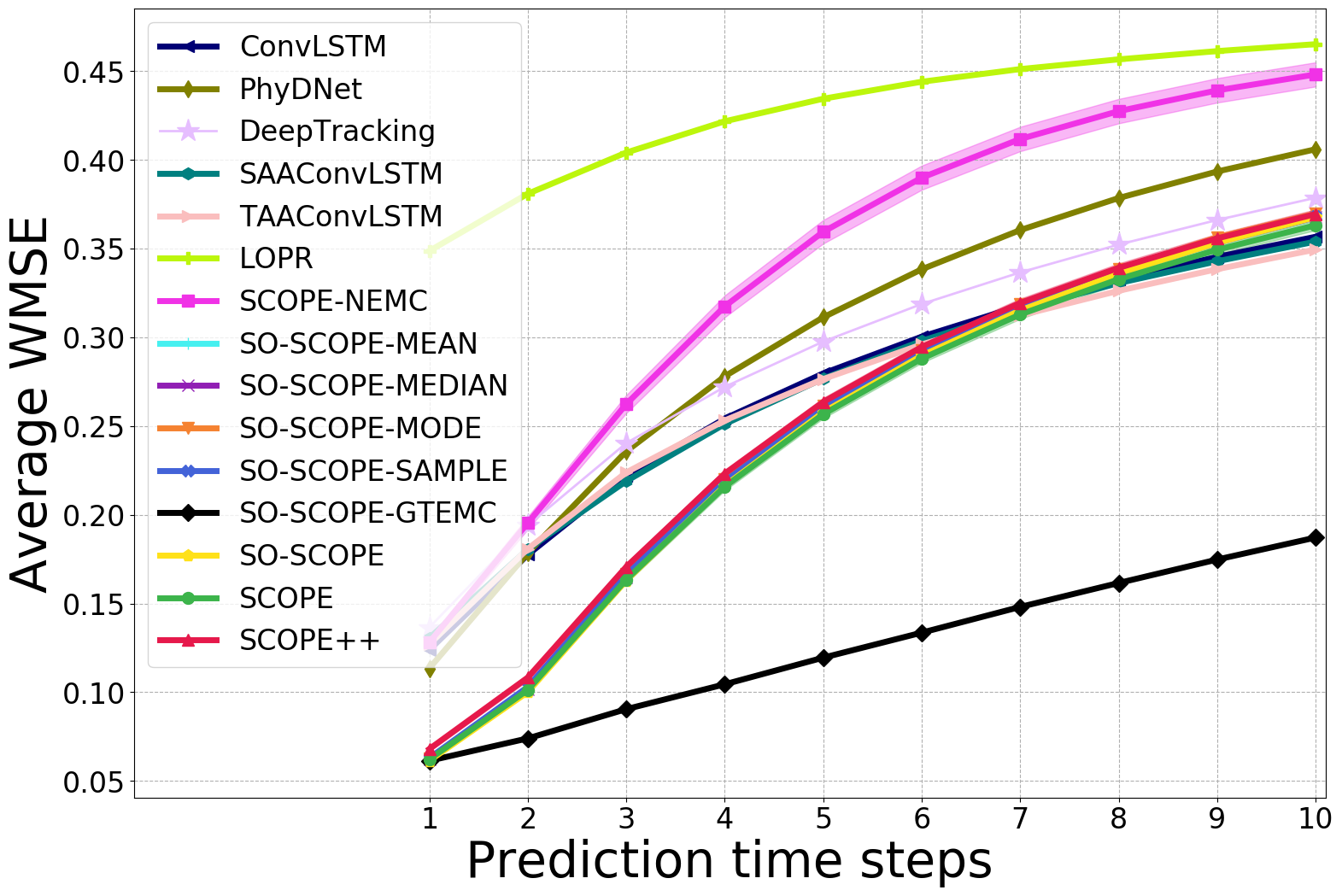}
            \label{fig:wmse_turtlebot2}
    }%
    \subfloat[WMSE: OGM-Jackal]{
            \centering
            \includegraphics[width=0.3\textwidth]{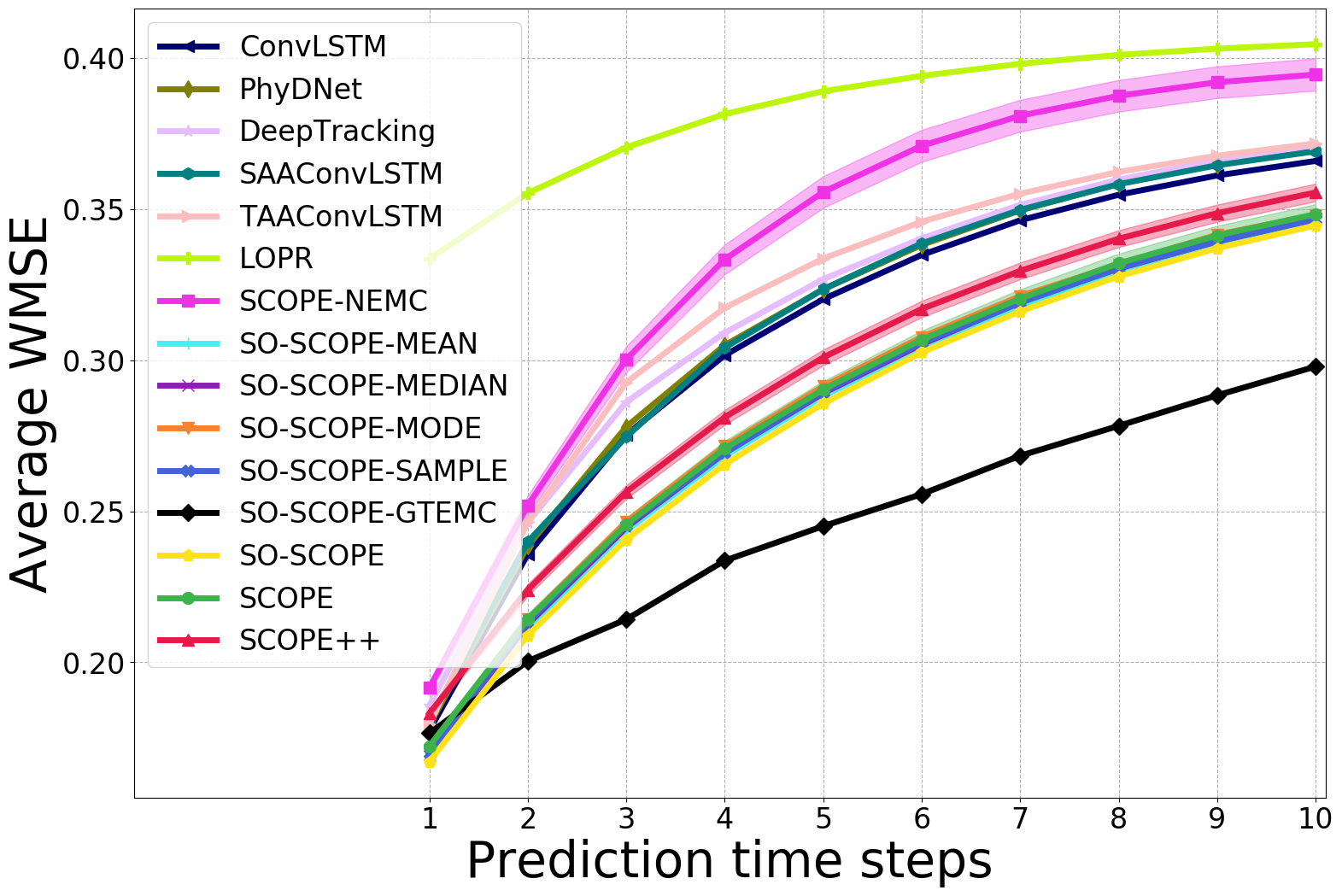}
            \label{fig:wmse_jackal}
    }%
    \subfloat[WMSE: OGM-Spot]{
            \centering
            \includegraphics[width=0.3\textwidth]{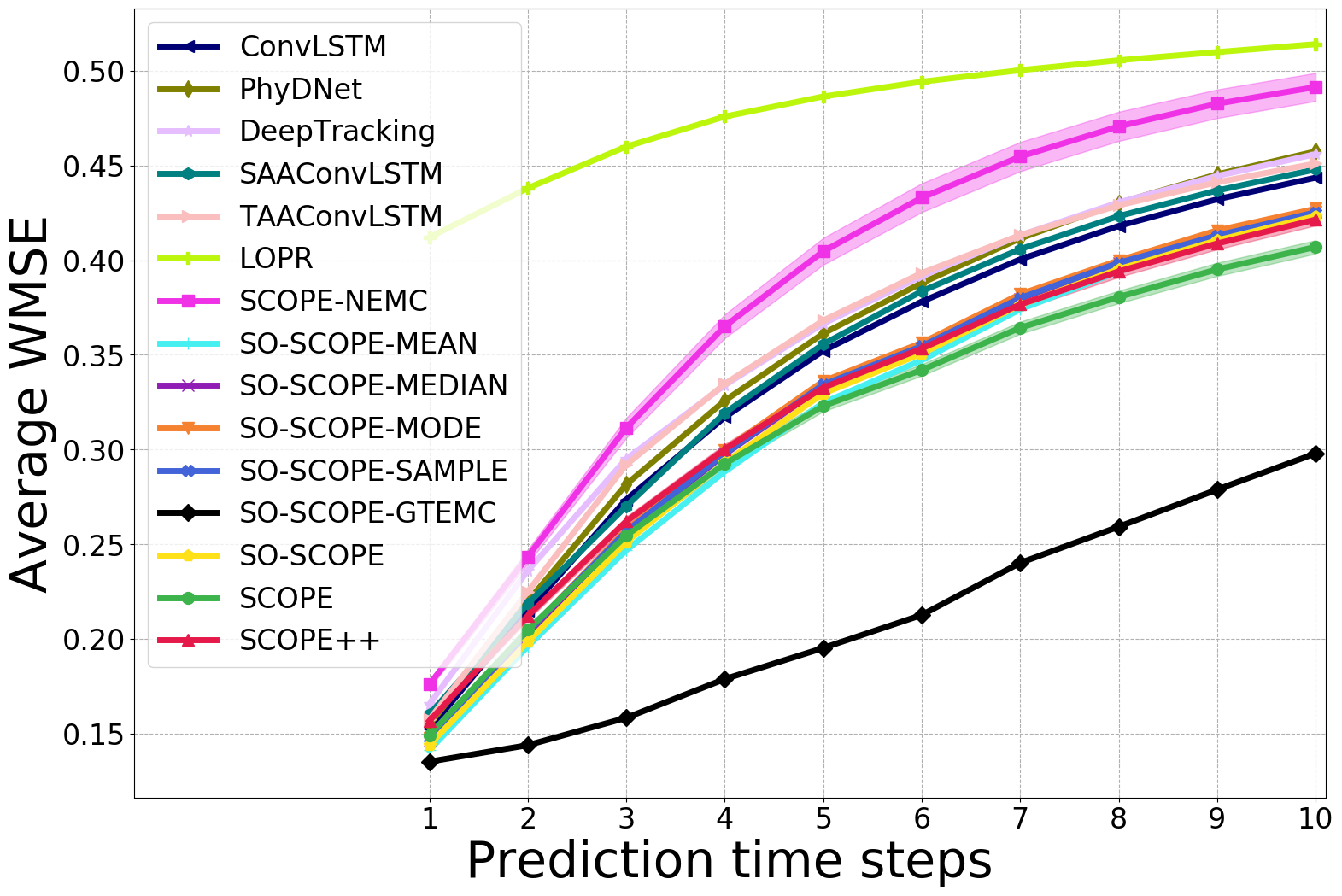}
            \label{fig:wmse_spot}
    }%
    \vspace{0.1pt}
    \subfloat[SSIM: OGM-Turtlebot2]{
            \centering
            \includegraphics[width=0.3\textwidth]{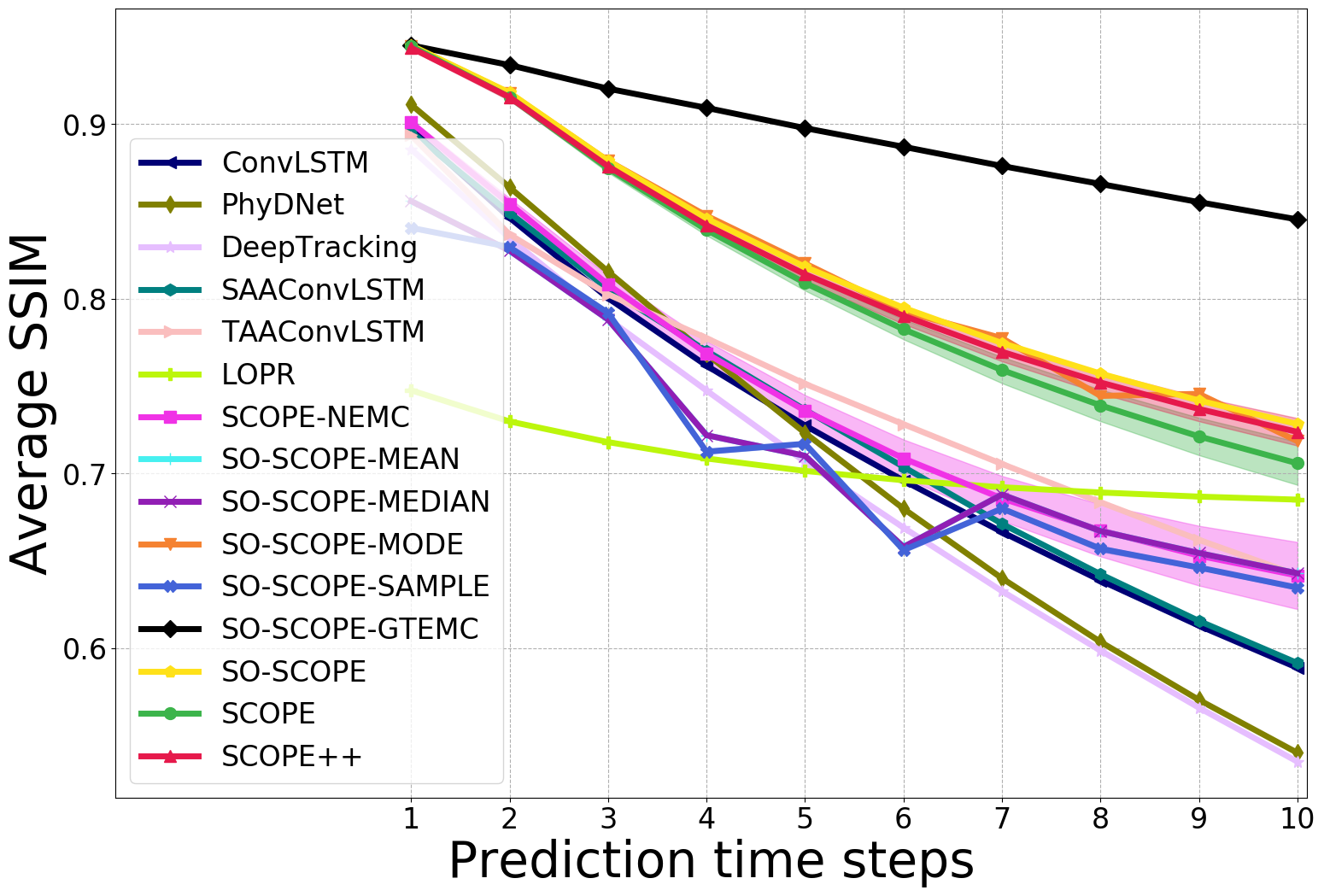}
            \label{fig:ssim_turtlebot2}
    }%
    \subfloat[SSIM: OGM-Jackal]{
            \centering
            \includegraphics[width=0.3\textwidth]{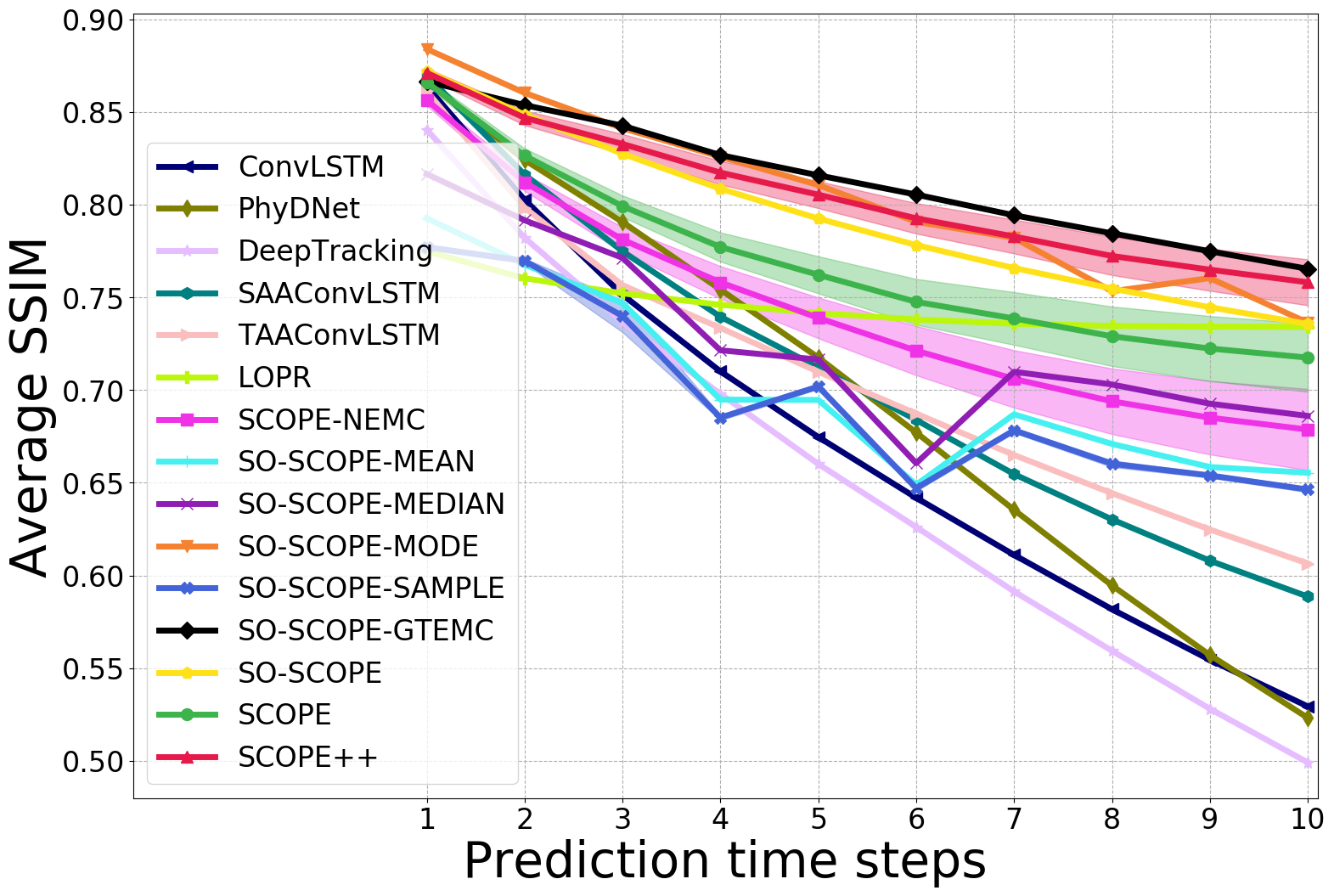}
            \label{fig:ssim_jackal}
    }%
    \subfloat[SSIM: OGM-Spot]{
            \centering
            \includegraphics[width=0.3\textwidth]{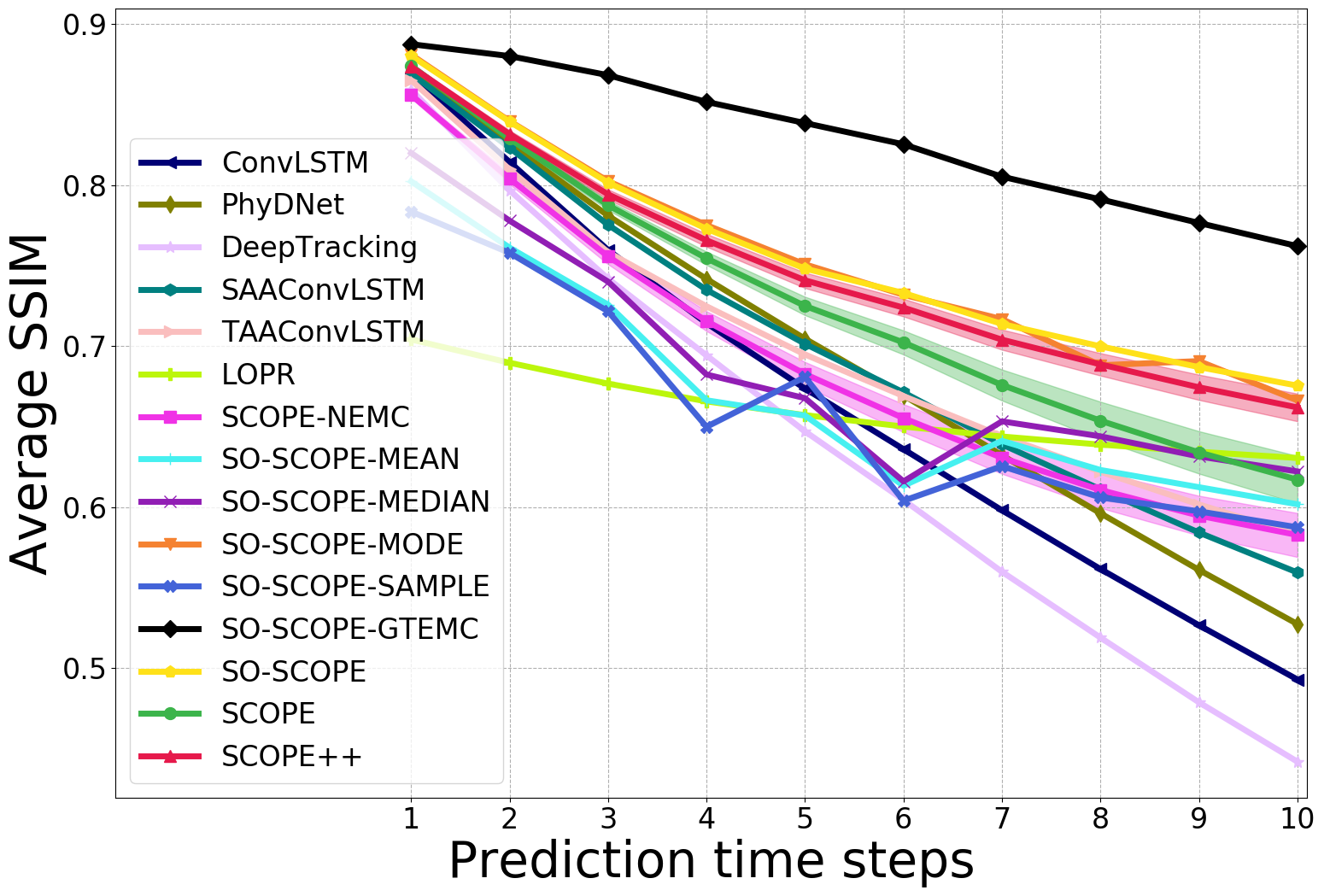}
            \label{fig:ssim_spot}
    }%
    \vspace{0.1pt}
    \subfloat[OSPA: OGM-Turtlebot2]{
            \centering
            \includegraphics[width=0.3\textwidth]{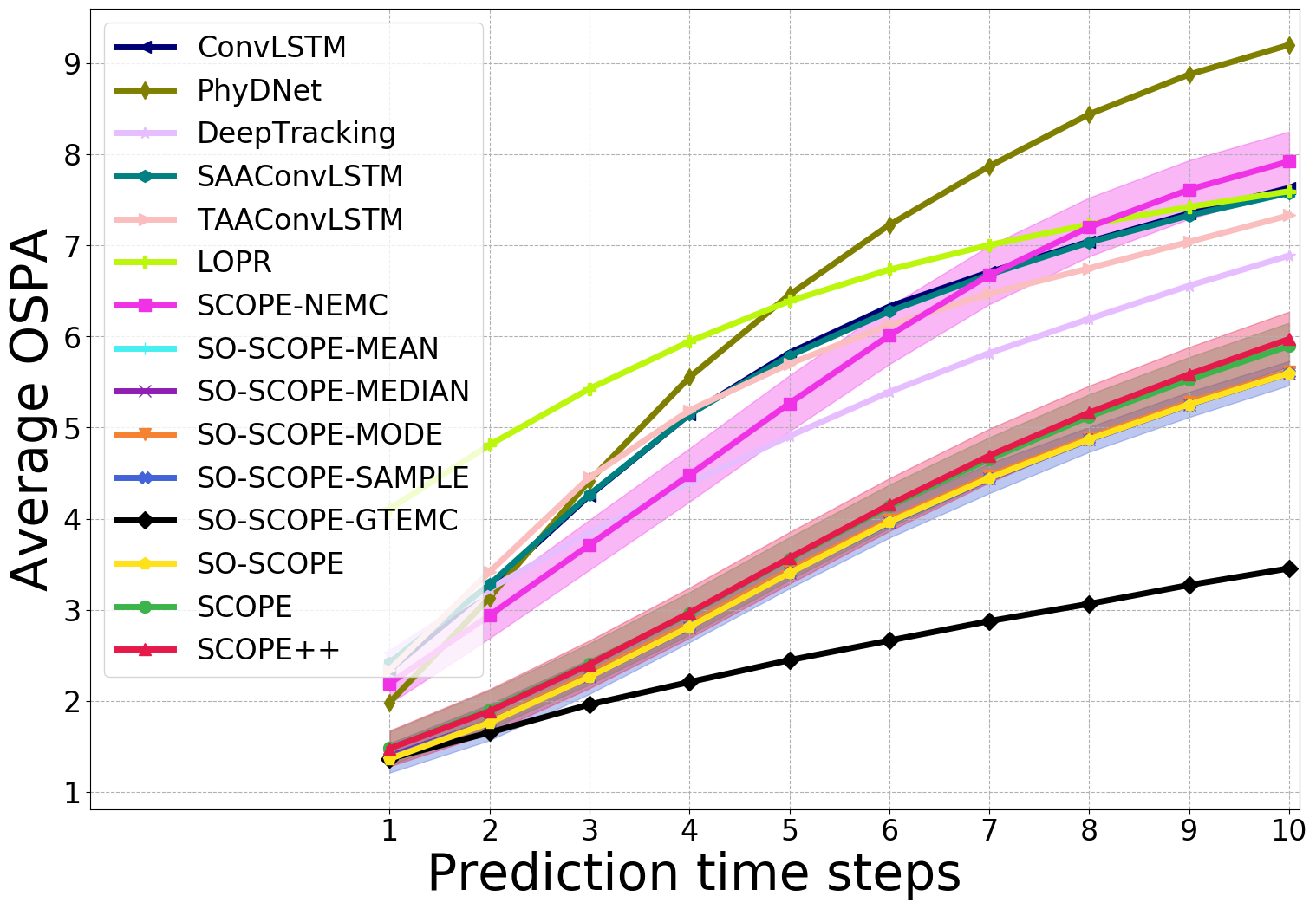}
            \label{fig:ospa_turtlebot2}
    }%
    \subfloat[OSPA: OGM-Jackal]{
            \centering
            \includegraphics[width=0.3\textwidth]{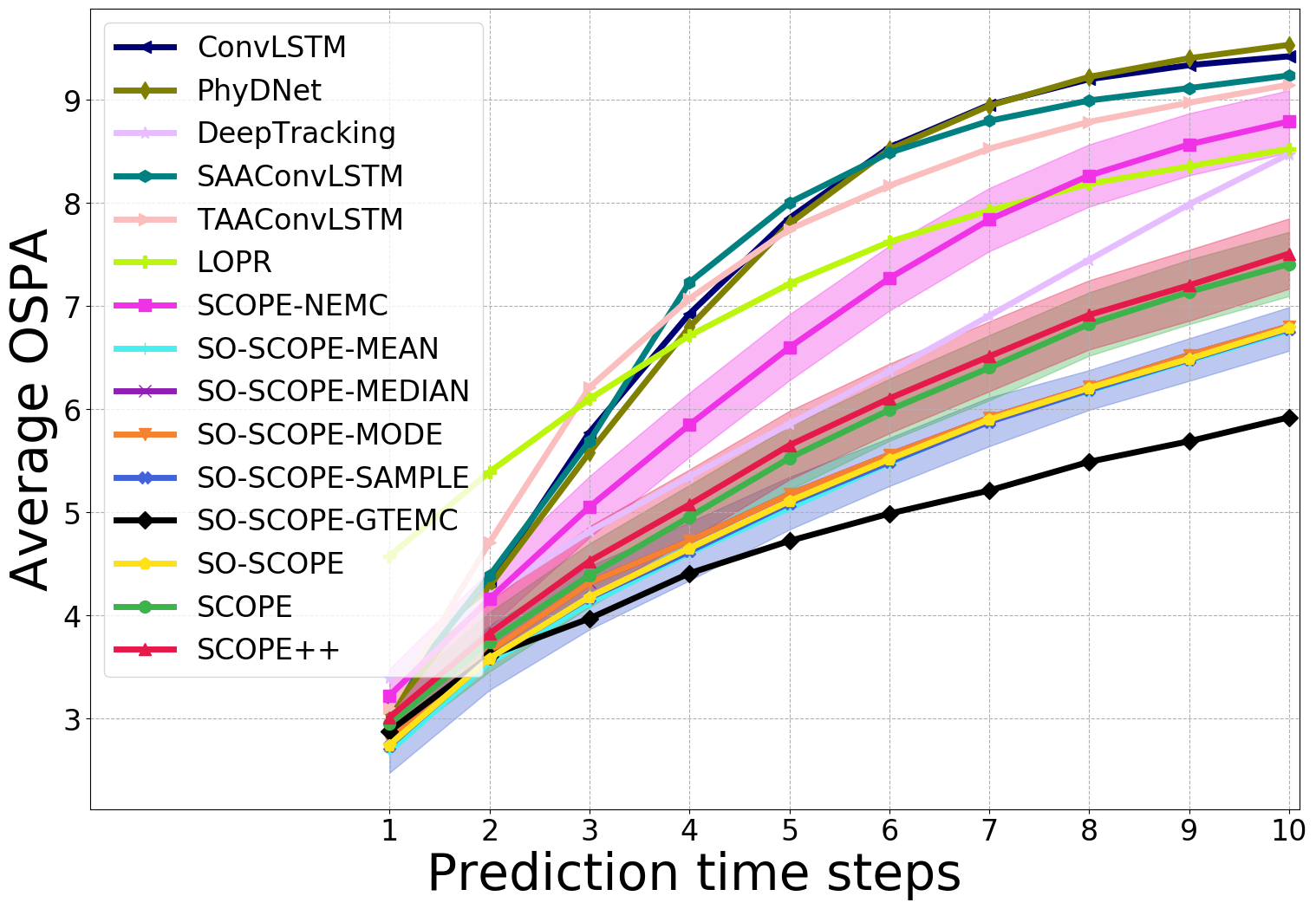}
            \label{fig:ospa_jackal}
    }%
    \subfloat[OSPA: OGM-Spot]{
            \centering
            \includegraphics[width=0.3\textwidth]{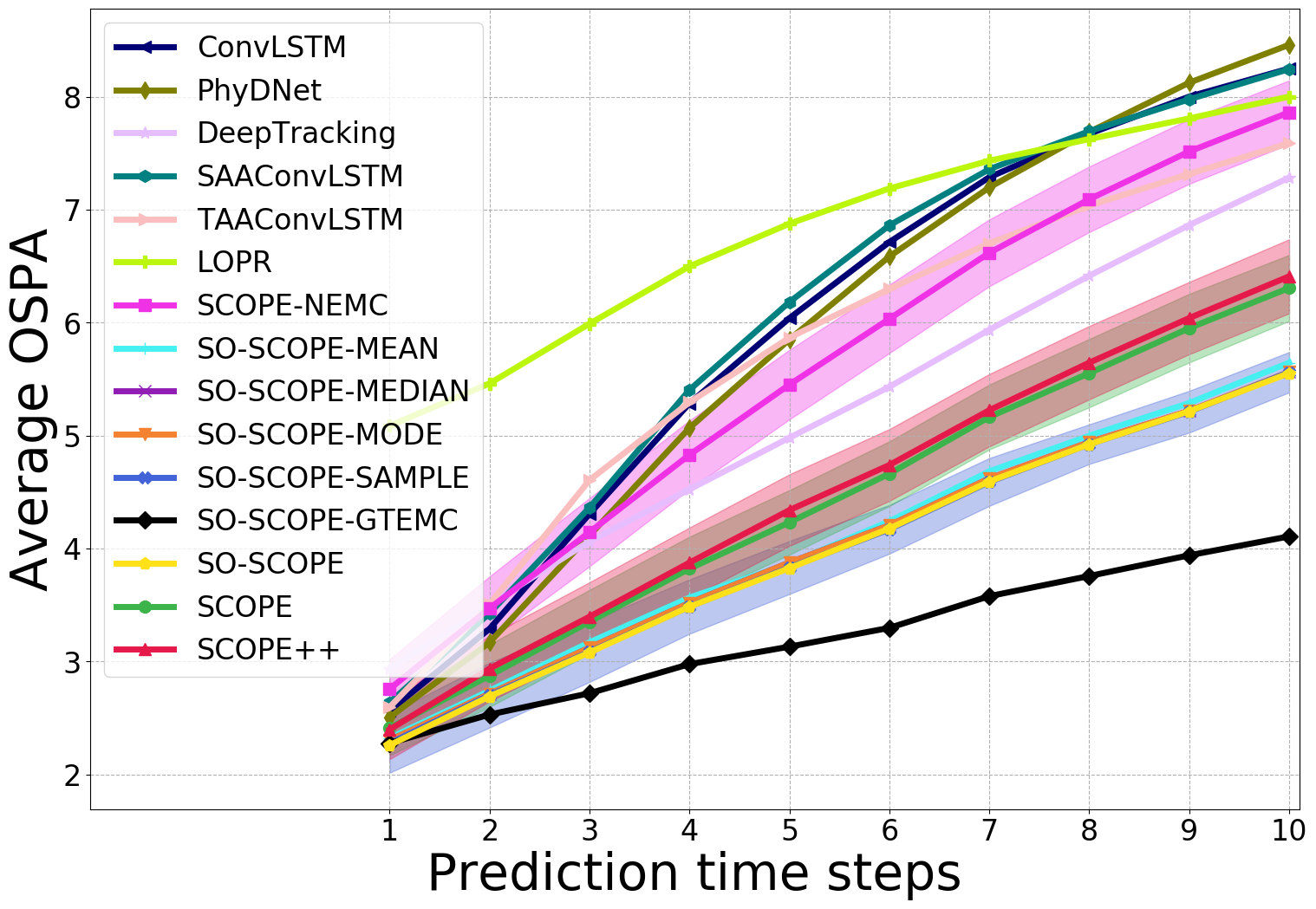}
            \label{fig:ospa_spot}
    }%
    \caption{WMSE (row 1), SSIM (row 2), and OSPA (row 3) for all tested methods on our 3 different datasets (columns).
    Each figure shows how the average value (over samples in the test datasets) of a given metric changes as a function of the prediction horizon, where lower is better for WMSE and OSPA and higher is better for SSIM.
    Lines for SCOPE-NEMC, SCOPE, and SCOPE++ include a 95\% confidence interval over 32 samples drawn from the VAE module.
    }
    \label{fig:prediction}
\end{figure*}

\paragraph{Absolute Error}
To evaluate the absolute error of predicted OGMs, we calculate the average WMSE of predicted OGMs for the next 10 prediction time steps, outlined in~\cref{fig:wmse_turtlebot2,fig:wmse_jackal,fig:wmse_spot}.
It can be seen that our proposed SCOPE family of methods with ego-motion compensation achieves significantly better average WMSE than the SCOPE-NEMC ablation baseline without ego-motion compensation in all test datasets, which illustrates the effectiveness of ego-motion compensation.
The average WMSEs of our proposed SCOPE family of methods at different prediction time steps are not significantly different from one another, but are all lower than the six image-based baselines (\ie ConvLSTM~\cite{shi2015convolutional}, DeepTracking~\cite{ondruska2016deep},  PhyDNet~\cite{guen2020disentangling}, SAAConvLSTM~\cite{lange2021attention}, TAAConvLSTM~\cite{lange2021attention}, and LOPR~\cite{lange2022lopr}).
This trend holds across all three test datasets collected by different robots.
This shows that the proposed SCOPE family of methods utilizing kinematic and dynamic information can predict more accurate OGMs than image-based approaches at different prediction time steps. 
Furthermore, the average WMSEs of the SO-SCOPE-GTEMC are much higher than that of the SO-SCOPE, which indicates that the best prediction performance can be achieved if our SO-SCOPE is equipped with an accurate customized ego-motion compensation algorithm for the robot instead of using only a general constant velocity motion model.

\paragraph{Structure Similarity}
To evaluate the structure similarity of predicted OGMs, we calculate the average SSIM of predicted OGMs for the next 10 prediction time steps, as shown in~\cref{fig:ssim_turtlebot2,fig:ssim_jackal,fig:ssim_spot}.
Unlike general WSME, which cares about the absolute error of a single OGM cell, SSIM cares about the entire OGM structure and its scene geometry.
First, we can see that the average SSIMs of our proposed SCOPE, SCOPE++, and SO-SCOPE predictors at different prediction time steps are significantly higher than that of SCOPE-NEMC, and the average SSIMs of SCOPE-NEMC are almost higher than the other six image-based baselines, which further illustrates the effectiveness of ego-motion compensation and also the advantage of our network design. 
Second, the average SSIMs of our software-optimized SO-SCOPE predictor and its statistical ablation baseline SO-SCOPE-MODE (\ie using peak values as cell prediction values) are not significantly different (similar to that of the SCOPE++ predictor), but much higher than the other three statistical ablation baselines (\ie SO-SCOPE-MEAN, SO-SCOPE-MEDIAN, SO-SCOPE-SAMPLE).
It shows that the direct output of SO-SCOPE ``student'' network can represent the prediction information and the mode from our prediction uncertainty statistics lookup table is also a good statistic to represent the prediction information (compared to other statistics like mean, median, and random variate), which further supports the choice of using knowledge distillation techniques and uncertainty quantification to software optimize our original SCOPE/SCOPE++ predictors.
Third, it is interesting that the average SSIM of our SCOPE++ with a local environment map is significantly higher than that of SCOPE without a local environment map in long-term predictions over multiple time steps.
This indicates that the local environment map that takes into account static objects helps to predict the long-term future of the environment. 
This is reasonable because the local environment map describes the basic scene geometry and shape.
Finally, the highest average SSIMs of SO-SCOPE-GTEMC also show the upper limit of the performance of our SO-SCOPE.

\begin{figure*}[t]
    \centering
    \subfloat[Inaccurate robot motion compensation]{
            \centering
            \includegraphics[width=0.45\textwidth]{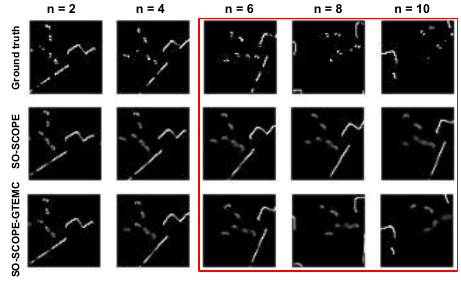}
            \label{fig:motion_failure}
    }%
    \subfloat[Occlusion or sudden appearance]{
            \centering
            \includegraphics[width=0.45\textwidth]{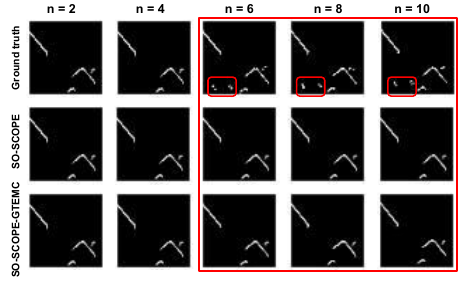}
            \label{fig:occlusion_failure}
    }%
    \caption{Two typical prediction failure cases of the SCOPE series predictors (SO-SCOPE as the example). The red bounding boxes highlight prediction errors.}
    \label{fig:failure}
\end{figure*}

\begin{table*}[t]
    \small\sf\centering
    \caption{Inference Speed, Model Size, and Memory Usage of different OGM predictors}
    \scalebox{0.72}{
        \begin{tabular}{c c c c c c c c c c}
            \toprule
            \textbf{Models} 
            & ConvLSTM~\cite{shi2015convolutional}  
            & PhyDNet~\cite{guen2020disentangling} 
            & DeepTracking~\cite{ondruska2016deep} 
            & SAAConvLSTM~\cite{lange2021attention}
            & TAAConvLSTM~\cite{lange2021attention}
            & LOPR~\cite{lange2022lopr}
            & SCOPE 
            & SCOPE++
            & SO-SCOPE\\
            \midrule
            \textbf{Inference Speed (FPS)} & 2.95 & 4.66 & 5.32 
            & 0.39 & 0.39 & 1.16
            & 23.29 & 10.68 & \textbf{34.75} \\
            \midrule
            \textbf{Model Size (MB)}  & 12.44  & 37.17  & \textbf{0.95}  
            & 21.64 & 22.02 & 1610.48 
            & 8.84  & 8.85  & 1.80  \\
            \midrule
            \textbf{Memory Usage (GB)}  & 0.70   &  0.71  & \textbf{0.63}   
            & 1.23  & 1.23  & 5.00 
            & 0.66  & 0.66  & 0.66  \\
           \bottomrule
        \end{tabular}
    }
    \label{tab:speed}
\end{table*}

\paragraph{Tracking Accuracy}
To further evaluate the tracking accuracy of predicted OGMs, we calculate the average OSPA of predicted OGMs for the next 10 prediction time steps, as shown in~\cref{fig:ospa_turtlebot2,fig:ospa_jackal,fig:ospa_spot}.
While the average WMSE and SSIM are the evaluation metrics from computer vision, the average OSPA is from the multi-target tracking and is more suitable for evaluating the physical quality of OGMs.
\Cref{fig:ospa_turtlebot2,fig:ospa_jackal,fig:ospa_spot} reveals that the average OSPA errors of our proposed SCOPE family of predictors are significantly lower than that of SCOPE-NEMC, and the average OSPA errors of SCOPE-NEMC are almost lower than that of other image-based baselines. 
This exciting result further demonstrates the preferential performance of our proposed motion-based methods (\ie SCOPE, SCOPE++, and SO-SCOPE) in tracking or predicting environmental states (\ie localization and cardinality) outperform SCOPE-NEMC, and while SCOPE-NEMC almost outperforms other image-based baselines (similar to the SSIM metric). 
Furthermore, the average OSPA errors of our software-optimized SO-SCOPE predictor and its statistical ablation baselines are not significantly different from one another (similar to the WMSE metric), but are lower than the original SCOPE and SCOPE++ predictors (especially in the real-world OGM-Jackal and OGM-Spot datasets), which is attributed to the use of knowledge distillation techniques to distill stochastic neural network models (\ie data augmentation).
It is reasonable and consistent with the argument in many knowledge distillation works~\cite{hinton2015distilling, stanton2021does} that the ``student'' network can generalize better than the ``teacher'' network.
However, the average OSPA errors of our proposed SCOPE and SCOPE++ predictors are almost the same, where the local environment map for static objects does not help SCOPE++ to reduce the OSPA error.
We believe that this is because the local environment map only provides useful information for static objects while the OSPA metric is biased towards dynamic objects (\eg pedestrians) since there tend to be more of them than static objects (\eg walls).
Finally, the highest average OSPAs of SO-SCOPE-GTEMC also show the upper limit of the performance of our SO-SCOPE. 

\subsection{Failure Case Analysis}
\label{subsubsec:failure_case_analysis}
Although our SCOPE series demonstrates its accurate and robust future state prediction ability, there are still two typical failure cases. 
The first failure case is when the robot moves erratically or rotates rapidly and the robot motion compensation is inaccurate, as shown in~\cref{fig:motion_failure}.
We can easily see that when the robot suddenly rotates at the 6th prediction time step, the SO-SCOPE predictor of our default constant velocity motion model fails to predict the later future states.
This is because our general constant velocity motion model fails to predict the robot's future motion, which is the key to OGM prediction.
To support our analysis, we also provide prediction results for the SO-SCOPE-GTEMC (which shows an upper bound on the performance of SO-SCOPE with ground truth motions).
From the last row of~\cref{fig:motion_failure}, we can see that if we can accurately compensate for the robot's motion, our SO-SCOPE can still provide accurate future predictions after the 6th prediction timestep.
Therefore, this failure issue can be easily alleviated by using more accurate robot motion models specific to the robots the researchers are using, as described in Sec.~\ref{subsec:robot_motion}.  

The second failure case is when the robot’s field of view (FOV) is blocked by a large obstacle or objects such as pedestrians suddenly appear in the robot’s FOV, as shown in~\cref{fig:occlusion_failure}.
We can see that when two pedestrians suddenly appear in the robot's FOV at the 6th prediction time step, our SO-SCOPE still misses the future states of these two pedestrians even with accurate motion compensation (\ie SO-SCOPE-GTEMC).
This is because there is not enough information about the object's motion to feed into the neural network predictor, which is also important for OGM prediction.
We believe that adding additional randomly sampled particles to the input OGMs or providing a longer OGM history can mitigate this occlusion or sudden appearance failure case.
Exploring the OGM prediction problem in occluded or sudden appearance scenes will be our future work.

\subsection{Computational Resource Utilization}
\label{subsec:computational_resource_utilization}
While it is encouraging that our proposed SCOPE family of OGM predictors exhibits good prediction performance, we also want to evaluate the inference speed, model size, and memory usage of all tested OGM predictors.
This is because mobile robots are resource-limited, and smaller model sizes, less memory usage, and faster inference speeds mean robots have a faster reaction time to face and handle dangerous situations in complex dynamic scenarios.

\subsubsection{Baseline Comparison}
\label{subsubsec:baseline_comparison}
\Cref{tab:speed} provides a comprehensive benchmark of resource usage, summarizing the inference speed, model size, and memory usage of nine predictors tested on a Jetson TX2 embedded computer equipped with a 256-core NVIDIA Pascal @ 1300MHz GPU and loaded with 8GB of memory.
We can see that: 
\begin{enumerate*}
    \item our SCOPE series achieves much faster inference than all other state-of-the-art baselines, especially our SO-SCOPE which can run at up to 35 FPS (about 89.1 times faster than the slowest SAAConvLSTM~\cite{lange2021attention}and TAAConvLSTM~\cite{lange2021attention}), 
    \item while our SCOPE and SCOPE++ models have the third-smallest model size, our proposed software-optimized SO-SCOPE model has the second-smallest model size and is approximately 894.7 times smaller than the largest LOPR~\cite{lange2022lopr} model, and
    \item our SCOPE series has the second smallest memory usage, which is about 7.6 times smaller than LOPR which has the largest memory usage of 5GB.
\end{enumerate*} 
These show that our SCOPE family can have real-time inference speed, reasonable model size, and small memory usage, which is more hardware-friendly than other state-of-the-art OGM predictors and can be deployed on resource-constrained robots.

\begin{figure}[t]
    \centering
    \includegraphics[width=0.45\textwidth]{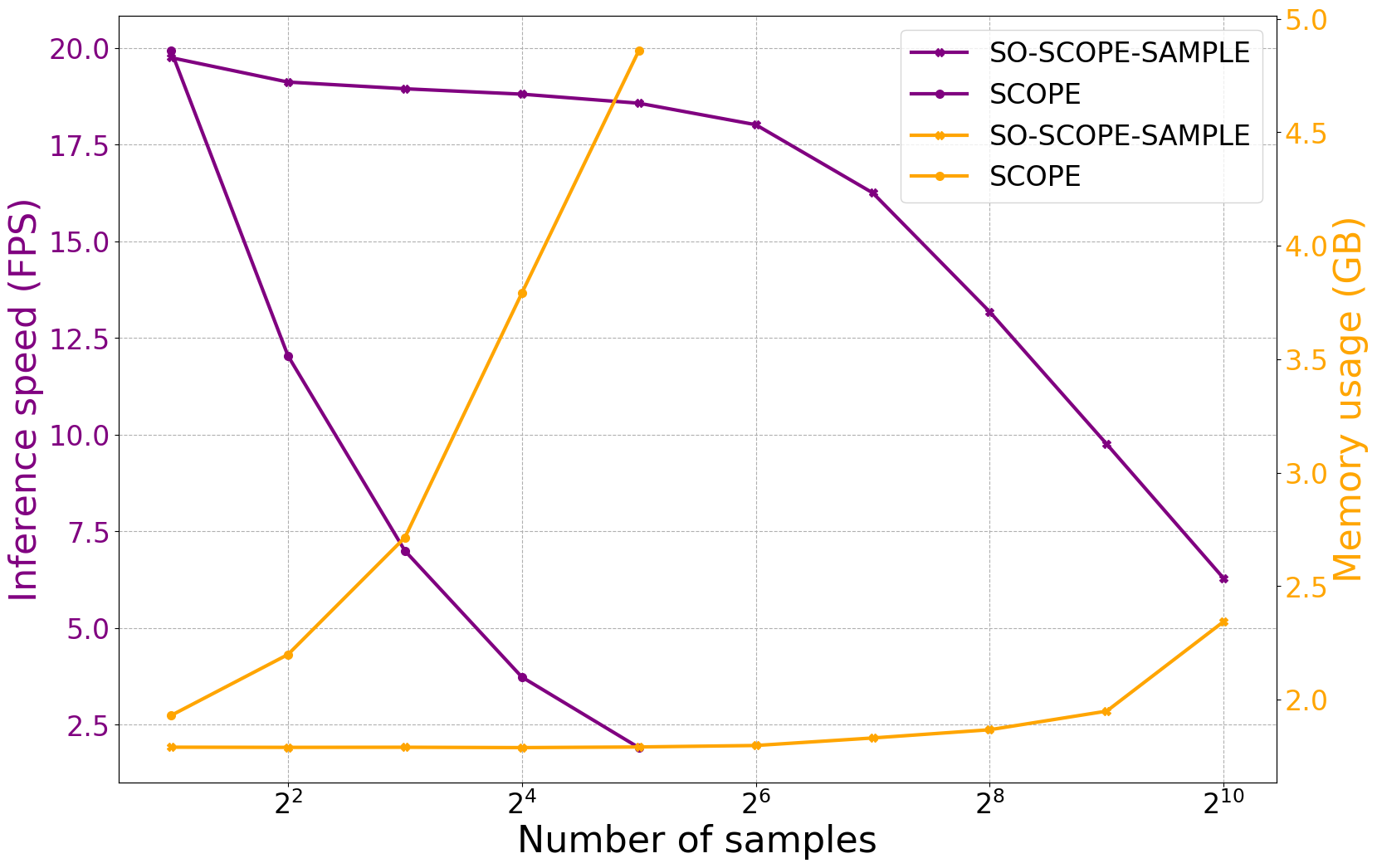}
    \caption{Inference speed and memory usage of SCOPE and SO-SCOPE predictors with a different number of samples.  
    }
    \label{fig:speed_vs_memory}
\end{figure}

\subsubsection{SCOPE Family Detailed Comparison}
\label{subsubsec:scope_family_detailed_comparison}
Then, we comprehensively compare the inference speed and memory usage of our software-optimized SO-SCOPE predictor with our SCOPE predictor (\ie a fast version of our SCOPE++) on a resource-constrained embedded computing device to show how our proposed software optimization approaches improve the embedded operation performance of SCOPE and SCOPE++.
For a fair comparison with our SCOPE predictor, we set the SO-SCOPE predictor into a mode that generates samples where the prediction uncertainty statistics lookup table will provide the random variables.
We compared their inference speed and memory usage when generating different numbers of samples, increasing as a power of 2 in the range $[2, 1024]$. 
\Cref{fig:speed_vs_memory} shows their detailed comparison results. 

As can be seen from \cref{fig:speed_vs_memory}, the inference speed of the SCOPE predictor drops significantly as the number of samples increases, with the lowest inference speed being about \unit[2]{FPS}.
In contrast, the inference speed of the SO-SCOPE hardly slows down as the number of samples increases until 128 samples are generated, and even with 1,024 samples it can still achieve around \unit[6]{FPS} (\ie the speed bottleneck is generating a large number of samples rather than network inference).  
This demonstrates the significant advantage of using knowledge distillation techniques to compress our SCOPE++ network, since smaller networks lead to faster inference speeds, and using prediction uncertainty statistics lookup table to provide uncertainty estimates for our SO-SCOPE predictor, which speeds up the time-consuming sample generation process.
Note that we could parallelize OGM sampling using our SO-SCOPE approach since it simply requires querying the lookup table, while SCOPE requires running the VAE model.

\Cref{fig:speed_vs_memory} also shows that the memory usage of the SCOPE predictor increases significantly with the number of samples and can only generate up to 32 samples on a Jetson TX2 device due to its \unit[8]{GB} memory limit.
In contrast, the memory usage of 
the SO-SCOPE predictor barely increases with the number of samples until 1,024 samples are generated (\ie memory bottleneck is storing a large number of samples).
It demonstrates the significant advantage of using the prediction uncertainty statistics lookup table to provide uncertainty estimates for our SO-SCOPE predictor in memory reduction.

\subsection{Uncertainty Characterization}
\label{subsec:uncertainty_characterization_results}
To characterize the uncertainty information of our SCOPE family of predictors, we use the Shannon entropy~\cite{shannon2001mathematical} to measure the uncertainty.
For an OGM ${\mathbf{o}}$, the Shannon entropy is given by $ H(\mathbf{o}) = -\sum_{c \in \mathbf{o}} p \log p + (1-p) \log (1-p)$,
where the first equality comes from the assumption that cells $c$ within an OGM are independent and the second equality comes from the definition of Shannon entropy for a Bernoulli distribution with parameter $p$ (since each cell can either be occupied, with probability $p$, or free, with probability $1-p$). 
In our case, we have a predicted map, $\hat{\mathbf{o}}$ and data about the distribution of occupancy values $\tilde{c}$, which comes from the predicted occupancy value $\hat{c}$ along with the mixture model \eqref{eq:mixture_model} and learned parameters $\mathbf{\xi}$.
Then, the entropy becomes
\begin{subequations}
\begin{align}
       H(\hat{\mathbf{o}}) &= \sum_{c \in \hat{\mathbf{o}}} E_{f_{\xi}(\hat{c})}[H(\tilde{c})], 
       \label{eq:entropy} \\
        = & -\sum_{c \in \hat{\mathbf{o}}} \int_0^1 \left[ \tilde{c} \log \tilde{c} + (1-\tilde{c}) \log (1 -\tilde{c}) \right] f_{\xi}(\tilde{c}) d \tilde{c}, 
        \label{eq:entropy_theory} \\
        \approx & - \sum_{c \in \hat{\mathbf{o}}} \frac{1}{M} \sum_{j=1}^{M}{\left[ \tilde{c}_{j} \log \tilde{c}_{j} + (1-\tilde{c}_{j}) \log (1 -\tilde{c}_{j}) \right]}, 
        \label{eq:entropy_estimation} 
\end{align}
\label{eq:entropy_calculation}
\end{subequations}
where $M$ is the number of predicted OGM samples and $\tilde{c}_{j}$ is the occupancy value of the cell in the $j$th predicted OGM $\tilde{\mathbf{o}}_j$.

Note that we can directly use \eqref{eq:entropy_theory} to calculate the predicted OGM entropy of SCOPE, SCOPE++, and SO-SCOPE because we have the distribution $f_{\xi}(c)$ for OGM cell $c$ from \eqref{eq:mixture_model}. 
We use the predicted mean map from OGM samples $\bar{\mathbf{o}} = \frac{1}{M}\sum_{j=1}^M \tilde{\mathbf{o}}$ instead of $\hat{\mathbf{o}}$ for SCOPE and SCOPE++, and the direct output of the SO-SCOPE network as $\hat{\mathbf{o}}$ for SO-SCOPE.
On the other hand, we also use \eqref{eq:entropy_estimation} to estimate the predicted OGM entropy of SCOPE and SCOPE++, since we can obtain samples of predicted OGM through the VAE module.
We also normalize the data by dividing the entropy by the total number of cells.

\subsubsection{Experiment Setup}
\label{subsubsec:experiment_setup}
Our characterization of the uncertainty has three primary goals:
\begin{enumerate*}
    \item show that the generated OGMs are consistent with reality,
    \item show that the SCOPE series converge to a consistent distribution, and
    \item show that the uncertainty in the overall OGM increases with the number of objects in the scene.
\end{enumerate*}    
For the experiments, we use the OGM-Turtlebot data dataset and fix the prediction horizon to $\tau=5$.
We use the evaluation pipeline used to calculate OSPA Error on OGMs in~\cite{xie2023sogmp} to count the number of objects in each OGM.
Then, we randomly select 20 input sequences for each number of objects (from 1 to 12) and generate a total of 1,024 OGM samples for each input sequence using each SCOPE method.

\subsubsection{Qualitative results}
\label{subsubsec:uncertianty_qualitative_results}
\begin{figure}[t]
    \centering
    \includegraphics[width=0.43\textwidth]{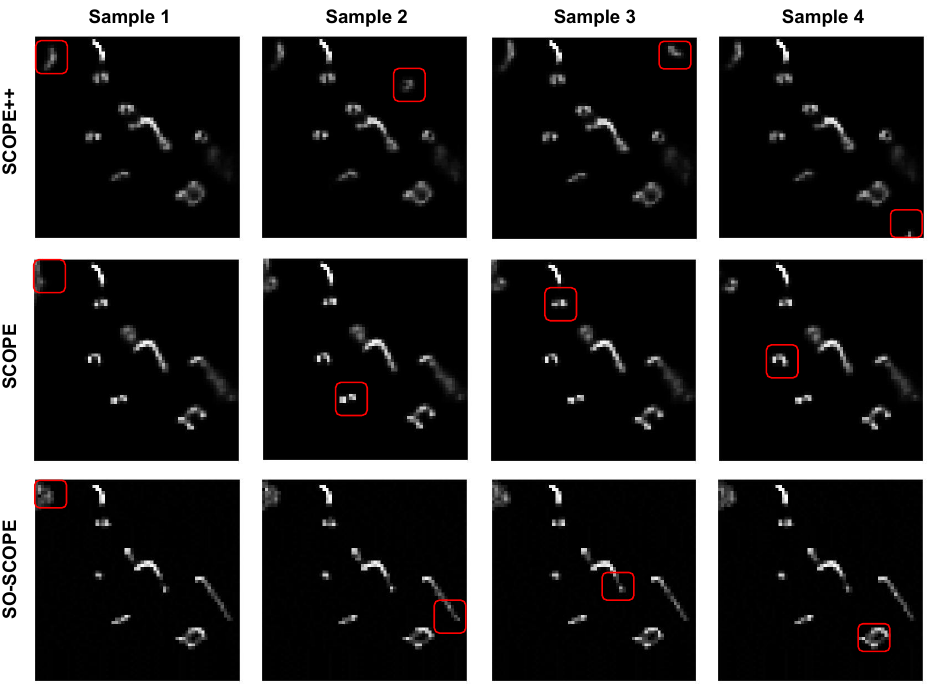}
    \caption{A diverse set of prediction samples of our SCOPE++, SCOPE, and SO-SCOPE predictors tested on the OGM-Turtlebot2 dataset at the 5th prediction timestep.
    The red bounding boxes highlight the regions of variation in the predicted samples for the same input in each stochastic algorithm (each row), which demonstrates the diversity of our proposed stochastic predictors.
    }
    \label{fig:showcase_samples}
\end{figure}

Our proposed SCOPE++, SCOPE, and SO-SCOPE have similar prediction performance and are able to generate prediction samples.
\Cref{fig:showcase_samples} showcases a diverse set of representative prediction samples.
We see that all three methods show variation across samples and generate realistic-looking structures, such as cylinders for legs and lines for walls.
We also see that SCOPE++ shows more variation across samples than SO-SCOPE and SCOPE.

\subsubsection{Quantitative results}
\label{subsubsec:uncertianty_quantitative_results}
One of the biggest differences between our proposed methods compared to state-of-the-art baselines is that the SCOPE family can provide uncertainty estimates.
The goal of these tests is to analyze the output distribution of our SCOPE, SCOPE++, and SO-SCOPE predictors.

\begin{figure}[t]
    \centering
    \subfloat[Entropy vs. Number of samples]{
            \centering
            \includegraphics[width=0.34\textwidth]{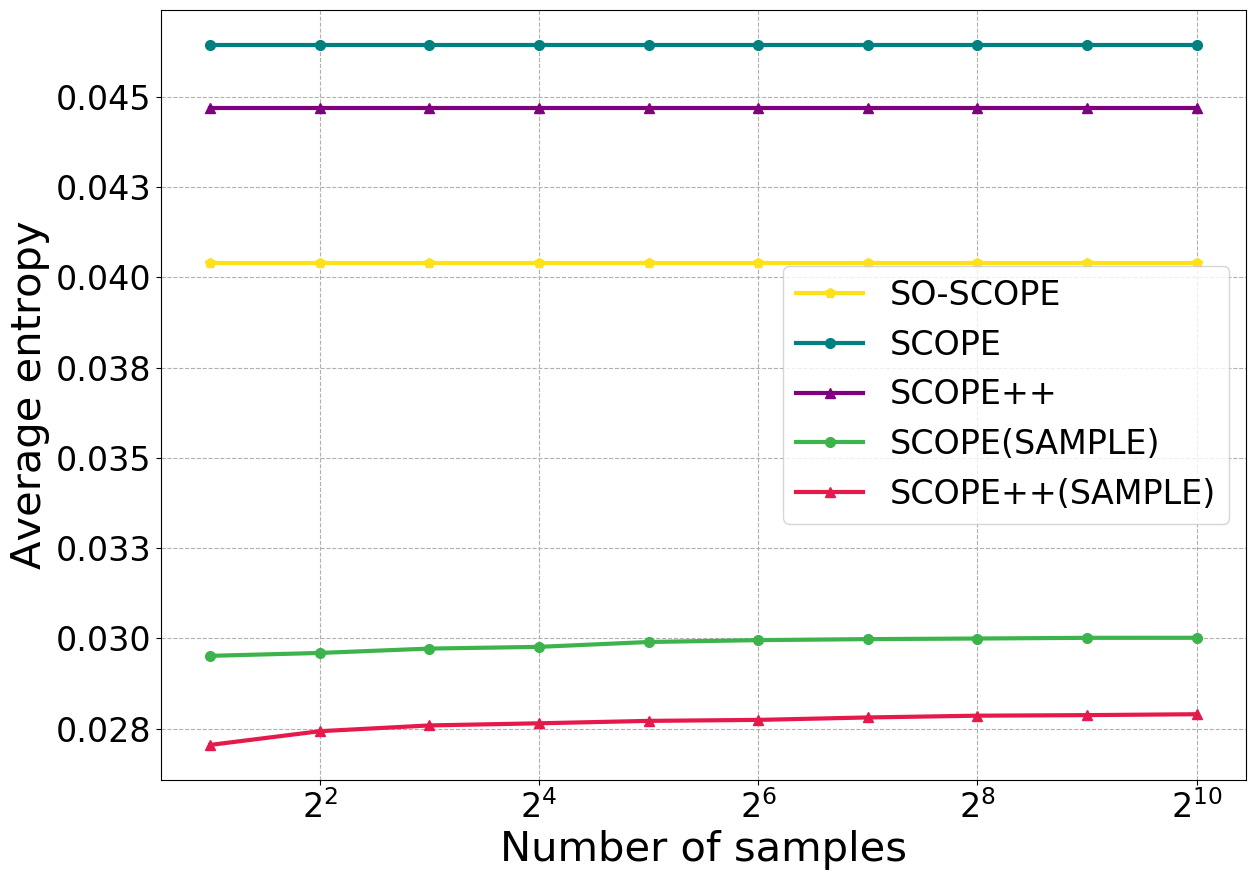}
            \label{fig:entropy_samples}
    }%
    \vspace{0.001cm}
    \subfloat[Entropy vs. Number of objects]{
            \centering
            \includegraphics[width=0.34\textwidth]{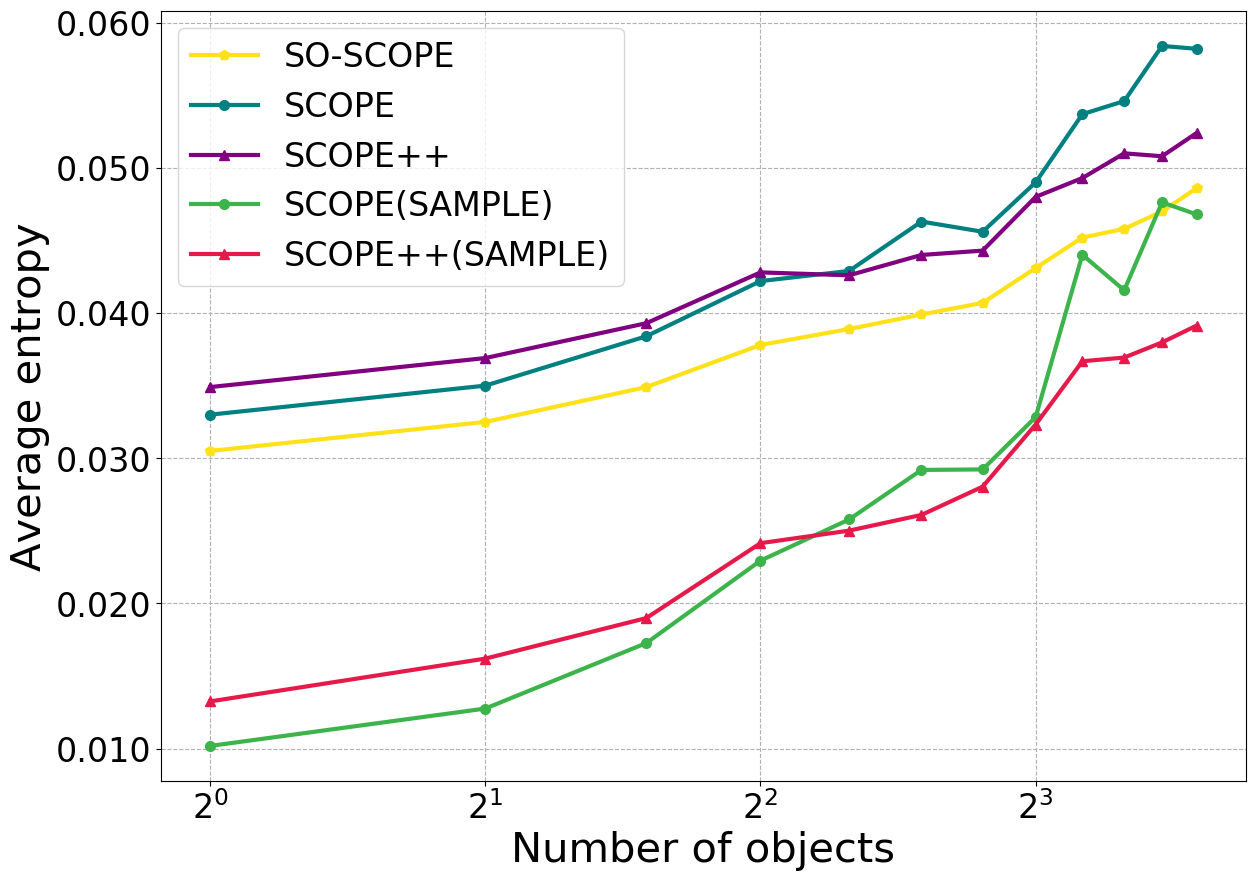}
            \label{fig:entropy_objects}
    }%
    \caption{Average entropy of our SCOPE, SCOPE++, and SO-SCOPE predictors at 5th prediction time step on OGM-Turtlebot2 dataset.
    }
    \label{fig:uncertainty}
\end{figure}

\begin{figure}[t]
    \centering
    \includegraphics[width=0.43\textwidth]{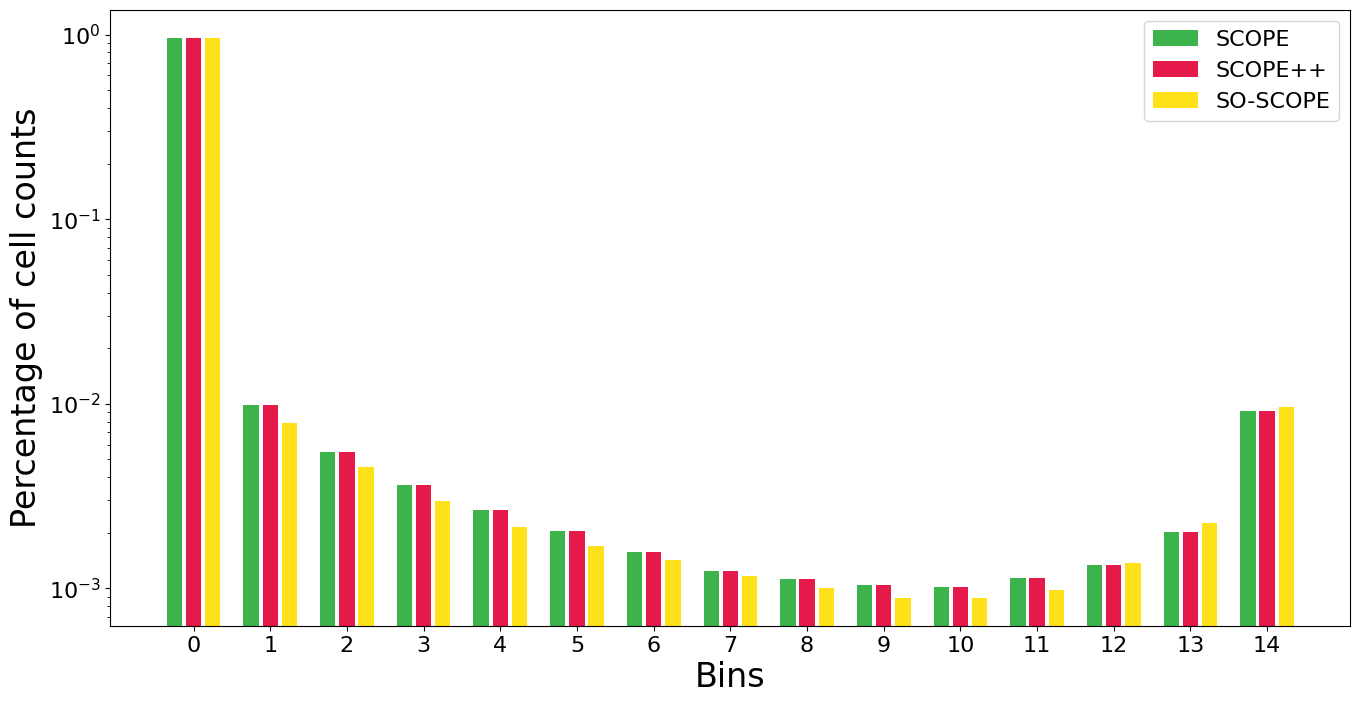}
    \caption{Cell counts for each bin in the SCOPE family of predictors. 
    }
    \label{fig:cell_counts}
\end{figure}

\paragraph{Convergence}
First, we show how the entropy of the final probabilistic OGM changes as the number of OGM samples increases, with the hypothesis that it will level off at some value well below that of the entropy of a uniform distribution. 
\Cref{fig:entropy_samples} shows that the average entropy of SCOPE(SAMPLE) and SCOPE++(SAMPLE) predictors (\ie computed via \eqref{eq:entropy_estimation}) starts to converge and plateau at a certain value after generating 128 samples, which is consistent with our hypothesis.
Note that the entropy values for SCOPE, SCOPE++, and SO-SCOPE are flat lines because we use \eqref{eq:entropy_theory} for calculation and no sampling is required.
Furthermore, the differences among the three flat lines are small (\ie no more than 0.004), which indicates that our uncertainty quantification for SCOPE++ is correct and can provide reasonable uncertainty estimates for our SO-SCOPE predictor.
The differences between them are mainly caused by the differences in their predicted OGMs.
\Cref{fig:cell_counts} shows the differences in cell counts for each bin in their predicted OGMs. 
It is consistent with the entropy differences of our SCOPE, SCOPE++, and SO-SCOPE predictors, where the difference between SCOPE and SCOPE++ is smaller than the difference between SO-SCOPE and SCOPE++ (\ie SO-SCOPE has a smaller percentage of cell counts or lower entropy near bin 1 to bin 11).

However, we can see that there is a gap between the entropy calculated by \eqref{eq:entropy_theory} using the mean map $\bar{\mathbf{o}}$ and the entropy calculated by \eqref{eq:entropy_estimation} (\ie SCOPE/SCOPE++ is higher than SCOPE(SAMPLE)/SCOPE++(SAMPLE)). 
We believe this gap stems from two factors.
First, our assumption that cells $c$ within an OGM are independent will overestimate the uncertainty values because actually knowing the occupancy of one cell should reduce the uncertainty of neighboring cells (and our learned latent representation in the VAE is encoding these correlations).
Second, the mean map $\bar{\mathbf{o}}$ is different from individual samples $\tilde{\mathbf{o}}$, so the statistics differ slightly.

\paragraph{Scene Complexity}
We also wish to show how the entropy of the final probabilistic OGM changes as the number of objects in the OGM increases, with the hypothesis that it will increase with the number of objects in it.
\Cref{fig:entropy_objects} clearly demonstrates that this is true, with the entropy of SCOPE/SCOPE(SAMPLE), SCOPE++/SCOPE++(SAMPLE) and SO-SCOPE predictors increasing with the number of objects.
This confirms the intuition that there is more uncertainty in a scene with more (dynamic) objects in it.


\begin{figure*}[t]
    \centering
    \includegraphics[width=0.8\textwidth]{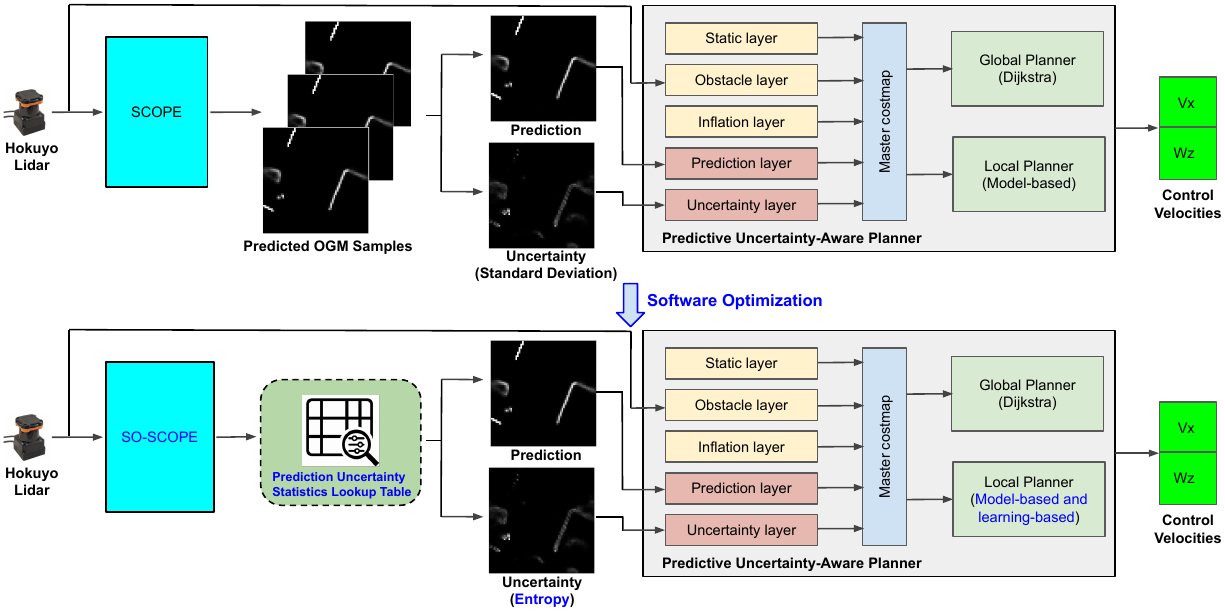}
    \caption{System architectures of the SCOPE-based and SO-SCOPE-based predictive uncertainty-aware navigation planners.
    The blue font emphasizes the difference between the SCOPE-based navigation framework and the SO-SCOPE-based navigation framework.
    The basic process of our proposed navigation framework is as follows: 
    first, the lidar data is also fed into our SCOPE or SO-SCOPE predictor to generate predicted OGM samples or lookup statistics from the prediction uncertainty statistics lookup table. Then, we can easily generate the prediction mean map and uncertainty map from these samples or the statistics. 
    Finally, we create the prediction costmap layer and the uncertainty costmap layer, combine them into the master costmap, and obtain our SCOPE-based or SO-SCOPE-based predictive uncertainty-aware planners.
    }
    \label{fig:planner}
\end{figure*}
\section{Uncertainty-Aware Navigation}
\label{sec:uncertainty_aware_navigation}
We next test the applicability of our SCOPE series to practical mobile robot navigation in crowded dynamic scenes.
To leverage the uncertainty information in the environmental future states and provide robust and reliable navigation behavior, we propose a general predictive uncertainty-aware navigation framework (shown in \cref{fig:planner}) based on costmaps.
Although this costmap-based framework loses the stochastic consistency of the entire navigation system, the biggest advantage is that it is used in the \texttt{move\_base} ROS navigation framework and can be integrated with most currently existing control policies. 
Specifically, we use prediction and uncertainty costmaps to tell the robot the potentially dangerous areas ahead of it and how certain they are, which enables mobile robots to make safer nominal path plans, take proactive actions to avoid potential collisions, and improve navigation capabilities.
To generate prediction and uncertainty costmaps from the output of our SCOPE family, we first binarize the predicted OGM and its uncertainty map using an occupancy threshold.
Second, we initialize a constant cost value for each binarized prediction and uncertainty map grid cell and generate the initial prediction and uncertainty costmaps.
Finally, we map each occupied grid cell of the prediction costmap and uncertainty costmap to a Gaussian obstacle value rather than a ``lethal'' obstacle value and obtain the final costmaps. 
This is because the predicted obstacles and uncertainty regions are not real obstacle spaces. 

There are two versions: based on SCOPE and SO-SCOPE.
The difference between SCOPE-based planner and SO-SCOPE-based planner is the method of generating prediction and uncertainty maps.
The SCOPE-based planner needs to use the VAE module to generate prediction samples to provide a prediction map (\ie mean) and an uncertainty map (\ie standard deviation), which requires a large number of samples to provide accurate estimates and is time-consuming and memory-consuming.
In contrast, our simple SO-SCOPE-based planner only needs to use a simple prediction uncertainty statistics lookup table to provide a prediction map (\ie SO-SCOPE output) and an uncertainty map (\ie entropy), which has fast running speed and small memory consumption.
Note that the SCOPE-based navigation framework can only run with traditional model-based local planners (\eg DWA~\cite{fox1997dynamic} and VO-based planners~\cite{fiorini1998motion, wilkie2009generalized}) in the resource-limited robots while the SO-SCOPE-based navigation framework can run with any types of model-based and learning-based local planners (\eg CNN~\cite{xie2021towards}, A1RC~\cite{guldenring2020learning}, and DRL-VO~\cite{xie2023drlvo}). 

\subsection{Baselines and Evaluation Metrics}
\label{subsubsec:baselines_and_evaluation_metrics_navigation}
Since our proposed predictive uncertainty-aware navigation framework can be combined with different types of existing control policies, we define the following naming convention for predictive uncertainty-aware control strategy instances: \texttt{[policy]/[predictor]/[P/PU]}, where \texttt{[policy]} is the control policy used, \texttt{[predictor]} is the OGM predictor used, \texttt{P} means using only the prediction map, and \texttt{PU} means using both the prediction map and its uncertainty map.  

Following this naming convention, we first use the model-based DWA~\cite{fox1997dynamic} as the local planner to instantiate a SCOPE-based predictive uncertainty-aware planner (\ie DWA/SCOPE/PU) to demonstrate how prediction and its uncertainty information improve robot navigation performance.
Then, we use the learning-based DRL-VO~\cite{xie2023drlvo} as the local planner to instantiate a SO-SCOPE-based predictive uncertainty-aware planner (\ie DRL-VO/SO-SCOPE/PU) to demonstrate the SO-SCOPE predictor is hardware friendly and our SO-SCOPE-based navigation framework can be applied to any currently existing control policies. 

We test these two predictive uncertainty-aware control policies, along with four state-of-the-art control policies: model-based DWA planner~\cite{fox1997dynamic}, supervised-learning-based CNN~\cite{xie2021towards}, DRL-based A1-RC~\cite{guldenring2020learning}, and DRL-based DRL-VO~\cite{xie2023drlvo}, and two ablation prediction-aware control policies without uncertainty maps: DWA/DeepTracking/P and DWA/SCOPE/P. 
Note that we use these four state-of-the-art baseline policies directly from their papers~\cite{fox1997dynamic, xie2021towards, guldenring2020learning, xie2023drlvo} without any retraining or parameter tuning. 
To evaluate the performance of navigation policies, we use the following four metrics from~\cite{xie2021towards, xie2023drlvo}: success rate, average time, average length, and average speed.

\subsection{Experiment Setup}
\label{subsubsec:experiment_setup_navigation}
\begin{figure}[t]
    \centering
    \subfloat[Gazebo lobby]{
            \centering
            \includegraphics[width=0.3\textwidth]{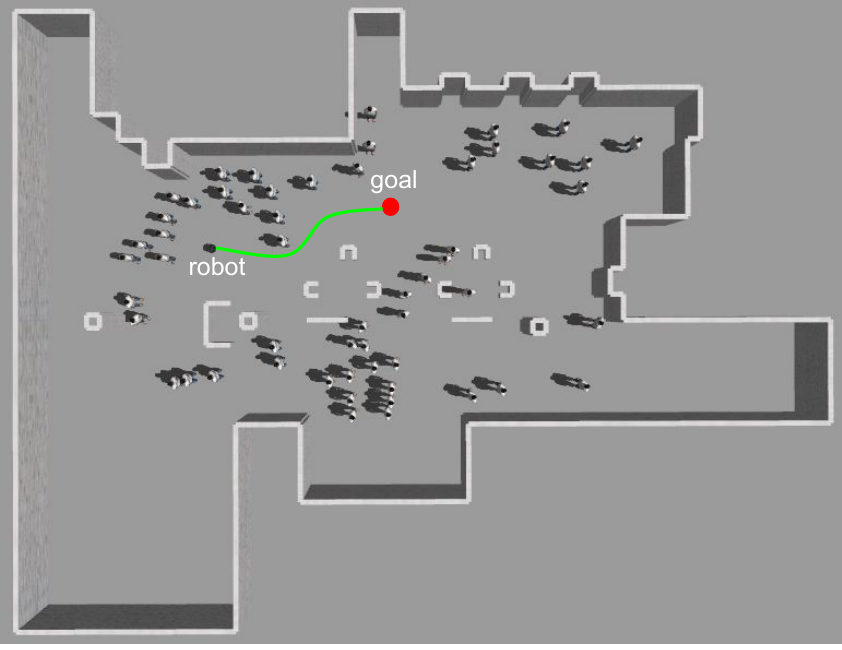}
            \label{fig:gazebo_lobby}
    }%
    \vspace{0.01cm}
    \subfloat[Indoor hallway]{
            \centering
            \includegraphics[width=0.344\textwidth]{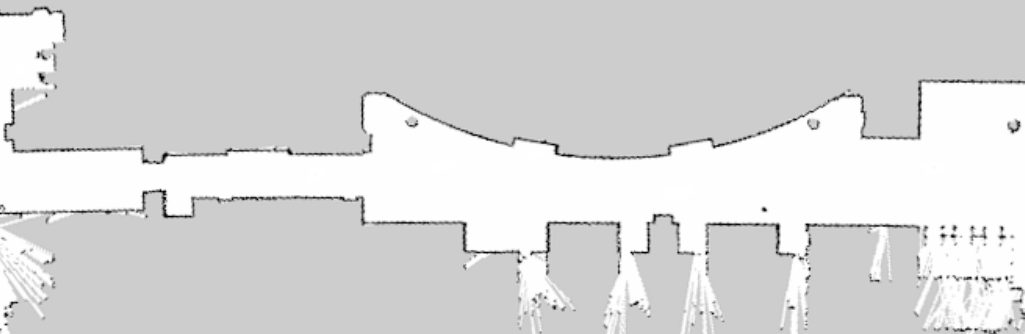}
            \label{fig:indoor_hallway}
    }%
 
    \caption{Robot navigation test environments in simulation and real-world. 
             The Lobby Gazebo world has 4 crowd density configurations ranging from 15-45 (sampling interval 10) pedestrians.
    }
    \label{fig:floorplan}
\end{figure}

Following the navigation evaluation settings in~\cite{xie2021towards,xie2023drlvo}, we conduct a total of 100 trials, and all control policies are tested by a Turtlebot2 robot with a maximum speed of \unit[0.5]{m/s}, equipped with a Hokuyo UTM-30LX lidar and a ZED stereo camera, in the simulated Lobby Gazebo world with 15-45 pedestrians (sampling interval 10), as shown in~\cref{fig:gazebo_lobby}.
The measurement range of Hokuyo lidar is set to \unit[{$[0.1, 30]$}]{m}, its FOV is $270^\circ$, and its angular resolution is $0.25^\circ$.
The depth range of ZED camera is set to \unit[{$[0.3, 20]$}]{m}, and its FOV is $90^\circ$.

In addition, we deploy our proposed DWA/SCOPE/PU and DRL-VO/SO-SCOPE/PU control policies on a real Turtlebot2 robot, which has the same configuration as the simulated robot but uses an NVIDIA Jetson AVG Xavier embedded computer as its main computing device. 
We conduct testing in an indoor hallway environment at Temple University during peak traffic periods between classes, the floor plan of which is shown in~\cref{fig:indoor_hallway}.
Note that considering the computational resources of Turtlebot2, we use the SCOPE predictor to only generate 8 predicted OGM samples at the 6th prediction time step (\ie \unit[0.6]{s}) for the DWA/SCOPE/PU control policy and it cannot combine with any learning-based control policies.
In contrast, our SO-SCOPE predictions do not suffer from such problems and can be combined with model-based or learning control strategies (\eg with DRL-VO in our instantiation).

\begin{table*}[t]
    \small\sf\centering
    \caption{Navigation results at different crowd densities}
    \scalebox{0.835}{
    \begin{tabular}{l l l l l l}
        \toprule
        \textbf{Environment} & \textbf{Method} & \textbf{Success Rate} & \textbf{Average Time (s)} & \textbf{Average Length (m)} & \textbf{Average Speed (m/s)} \\ 
        \midrule

        \multirow{6}{*}{\begin{tabular}[c]{@{}c@{}}Lobby world, \\ 15 pedestrians\end{tabular}} 
         & DWA \cite{fox1997dynamic} & 0.94 & 12.21 & 5.09 & 0.42  \\  
         & CNN \cite{xie2021towards} & - & - & - & -   \\  
         & A1-RC \cite{guldenring2020learning} & 0.94 & 14.36 & 6.40 & 0.45    \\ 
         & DRL-VO \cite{xie2023drlvo} & 0.95 & \textbf{11.34} & 5.26 & \textbf{0.46} \\
         \cmidrule(lr){2-6}  
         
         & DWA/DeepTracking/P & 0.95 & 12.10 & \textbf{5.03} & 0.42  \\  
         & DWA/SCOPE/P & 0.93 & 12.62 & 5.06 & 0.40  \\  
         & DWA/SCOPE/PU & \textbf{0.98} & 12.79 & 5.06 & 0.40  \\ 
         \cmidrule(lr){2-6}  
         
         & DWA/SO-SCOPE/P & 0.94 & 12.12 & 5.10 & 0.42  \\  
         & DWA/SO-SCOPE/PU & 0.96 & 13.25 & 5.12 & 0.39  \\  
         & DRL-VO/SO-SCOPE/PU & 0.97 & 11.69 & 5.43 & \textbf{0.46} \\
         \midrule

        \multirow{6}{*}{\begin{tabular}[c]{@{}c@{}}Lobby world,\\ 25 pedestrians\end{tabular}} 
         & DWA \cite{fox1997dynamic} & 0.82 & 13.49 & 5.12 & 0.38  \\ 
         & CNN \cite{xie2021towards} & 0.80 & 19.31 & 6.16 & 0.32  \\  
         & A1-RC \cite{guldenring2020learning} & 0.88 & 14.18 & 6.26 & 0.44  \\
         & DRL-VO \cite{xie2023drlvo} & 0.92 & \textbf{11.37} & 5.29 & \textbf{0.47}  \\
         \cmidrule(lr){2-6}  
          
         & DWA/DeepTracking/P & 0.87 & 12.67 & \textbf{5.06} & 0.40  \\  
         & DWA/SCOPE/P & 0.90 & 13.53 & 5.08 & 0.37  \\  
         & DWA/SCOPE/PU & 0.91 & 13.75 & 5.07 & 0.37  \\
         \cmidrule(lr){2-6}  
          
         & DWA/SO-SCOPE/P & 0.90 & 13.23 & 5.13 & 0.39  \\ 
         & DWA/SO-SCOPE/PU & 0.92 & 14.26 & 5.19 & 0.36  \\  
         & DRL-VO/SO-SCOPE/PU & \textbf{0.93} & 11.95 & 5.52 & {0.46} \\
         \midrule
        
        \multirow{6}{*}{\begin{tabular}[c]{@{}c@{}}Lobby world, \\ 35 pedestrians\end{tabular}} 
         & DWA \cite{fox1997dynamic} & 0.82 & 14.18 & 5.15 & 0.36  \\
         & CNN \cite{xie2021towards} & 0.81 & 14.30 & 5.40 & 0.38  \\  
         & A1-RC \cite{guldenring2020learning} & 0.77 & 16.81 & 6.89 & 0.41  \\   
         & DRL-VO \cite{xie2023drlvo} & 0.88 & \textbf{11.42} & 5.31 & \textbf{0.46}  \\
         \cmidrule(lr){2-6}  
         
         & DWA/DeepTracking/P & 0.84 & 13.93 & \textbf{5.10} & 0.37  \\  
         & DWA/SCOPE/P & 0.86 & 13.79 & 5.12 & 0.34  \\  
         & DWA/SCOPE/PU  & 0.89 & 14.90 & 5.12 & 0.34  \\
         \cmidrule(lr){2-6}  
         
         & DWA/SO-SCOPE/P & 0.86 & 14.03 & 5.16 & 0.37  \\ 
         & DWA/SO-SCOPE/PU & 0.88 & 15.53 & 5.20 & 0.33  \\  
         & DRL-VO/SO-SCOPE/PU & \textbf{0.92} & 12.26 & 5.57 & 0.45 \\
         \midrule
         
         \multirow{6}{*}{\begin{tabular}[c]{@{}c@{}}Lobby world, \\ 45 pedestrians\end{tabular}} 
         & DWA \cite{fox1997dynamic} & 0.77 & 15.39 & 5.16 & 0.34  \\
         & CNN \cite{xie2021towards} & 0.79 & 16.65 & 5.62 & 0.34  \\   
         & A1-RC \cite{guldenring2020learning} & 0.77 & 14.65 & 6.28 & 0.43  \\ 
         & DRL-VO \cite{xie2023drlvo} & 0.81 & \textbf{11.65} & 5.37 & \textbf{0.46}  \\
         \cmidrule(lr){2-6}  
         
         & DWA/DeepTracking/P & 0.78 & 15.23 & \textbf{5.14} & 0.34  \\  
         & DWA/SCOPE/P & 0.79 & 14.84 & \textbf{5.14} & 0.35  \\  
         & DWA/SCOPE/PU  & 0.82 & 15.96 & 5.17 & 0.32  \\
         \cmidrule(lr){2-6}  
         
         & DWA/SO-SCOPE/P & 0.78 & 14.89 & 5.21 & 0.35  \\ 
         & DWA/SO-SCOPE/PU & 0.80 & 16.29 & 5.27 & 0.32  \\  
         & DRL-VO/SO-SCOPE/PU & \textbf{0.83} & 12.24 & 5.56 & 0.45 \\
         \bottomrule
    \end{tabular}
    }
    \label{tab:densities}
\end{table*}

\subsection{Simulation Results}
\label{subsubsec:simulation_results}
\Cref{tab:densities} summarizes these navigation results, where we observe three key phenomena.
First, compared to DWA series variants (\ie DWA, DWA/DeepTracking/P, DWA/SCOPE/P, and DWA/SO-SCOPE/P), our proposed DWA/SCOPE/PU and DWA/SO-SCOPE/PU policies have the highest success rate in each crowd size, while DWA/SCOPE/PU has almost the shortest path length.
This shows that the prediction costmap from our OGM predictors is able to help the currently existing model-based navigation policy (\ie DWA) to provide safer and shorter paths, and combining it with its associated uncertainty costmap can achieve a much better navigation performance (\ie PU is better than P alone).
Furthermore, it shows that our software-optimized SO-SCOPE predictor can maintain good robot navigation performance even though it leads to longer paths compared to the SCOPE predictor.
All these results demonstrate that our proposed SCOPE series predictors can improve safe robot navigation in crowded dynamic scenes.

\begin{figure*}[t]
    \centering
    \subfloat[DWA~\cite{fox1997dynamic}]{
            \centering
            \includegraphics[width=0.235\textwidth]{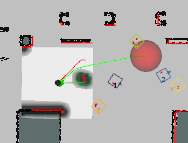}
            \label{fig:dwa}
    }%
    \subfloat[DWA/SCOPE/P]{
            \centering
            \includegraphics[width=0.235\textwidth]{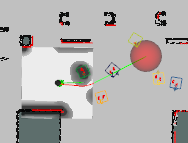}
            \label{fig:dwa_p}
    }%
    \subfloat[DWA/SCOPE/PU]{
            \centering
            \includegraphics[width=0.235\textwidth]{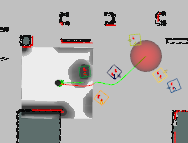}
            \label{fig:dwa_pu}
    }%
    \subfloat[DWA/SO-SCOPE/PU]{
            \centering
            \includegraphics[width=0.2352\textwidth]{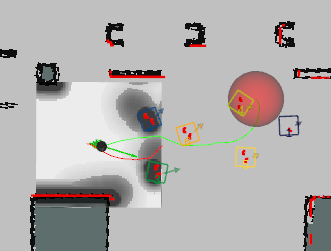}
            \label{fig:dwa_so_pu}
    }%
    \caption{Robot reactions and their final costmaps generated by different DWA-based control policies in the simulated lobby environment, which shows that the final master costmap generated by the prediction map and its uncertainty map can provide safer path planning.
    The robot (black disk) avoids pedestrians (colorful square boxes, each color represents a pedestrian) and reaches the goal (red disk) according to the nominal path (green curve) and local path (red curve) planned by the costmap (square grey map).
    }
    \label{fig:nav_costmaps}
\end{figure*}

The reasons why the OGM prediction and its uncertainty information can improve robot safe navigation can be explained qualitatively through~\cref{fig:nav_costmaps}.
It shows the difference of nominal paths and costmaps generated by four planners (\ie DWA, DWA/SCOPE/P, DWA/SCOPE/PU, and DWA/SO-SCOPE/PU) in the simulated lobby environment. 
The default DWA planner~\cite{fox1997dynamic} only cares about the current state of the environment and generates a costmap based on the perceived obstacles.
The predictive DWA planner (\ie  DWA/SCOPE/P) using the prediction map of our proposed SCOPE predictor can generate a costmap with predicted obstacles.
The predictive uncertainty-aware DWA planner (\ie DWA/SCOPE/PU and DWA/SO-SCOPE/PU) using both the prediction map and uncertainty map of our proposed SCOPE predictor can generate a safer costmap with predicted obstacles and uncertainty regions. 
Note that the path generated by DWA/SO-SCOPE/PU is more tortuous than that of DWA/SCOPE/PU, which may explain why DWA/SO-SCOPE/PU requires longer paths to achieve a similar success rate as DWA/SCOPE/PU.
These additional predicted obstacles and uncertainty regions of our proposed DWA/SCOPE/PU and DWA/SO-SCOPE/PU planners enable the robot to follow safer nominal paths and reduce collisions with obstacles, especially moving pedestrians.
See the accompanying Multimedia for a detailed simulation navigation demonstration.

Second, our proposed DWA/SCOPE/P and DWA/SO-SCOPE/P have a higher success rate than DWA/DeepTracking/P at every crowd size, indicating that higher prediction accuracy can also improve safe navigation performance. 
Third, compared with all other state-of-the-art control policies, our proposed DRL-VO/SO-SCOPE/PU policy has the highest success rate and the fastest average speed in every situation.
This suggests that our software-optimized SO-SCOPE predictor using a simple knowledge distillation network and prediction uncertainty statistical lookup table can also provide additional benefits for the learning-based safe navigation performance in crowded dynamic environments with varying crowd densities.
Furthermore, it shows the potential capability of our software-optimized predictive uncertainty-aware navigation framework for different high-computational load learning-based policies.

\begin{figure*}[t]
    \centering
    \subfloat[t]{
            \centering
            \includegraphics[width=0.47\textwidth]{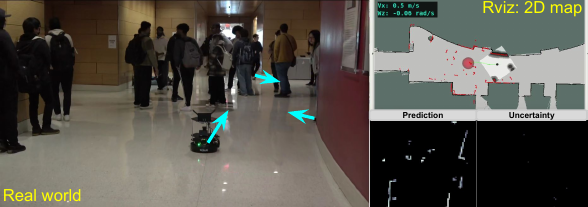}
            \label{fig:hallway_dwa_scope3}
    }%
    \subfloat[t+3]{
            \centering
            \includegraphics[width=0.47\textwidth]{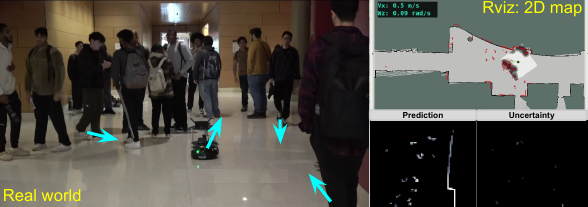}
            \label{fig:hallway_dwa_scope4}
    }%
    
    \caption{Robot deployed with DWA/SCOPE/PU reactions to moving pedestrians in the indoor hallway with the high crowd density at different times.
            }
    \label{fig:hallway_dwa}
\end{figure*}

\begin{figure*}[t]
    \centering
     \subfloat[t]{
            \centering
            \includegraphics[width=0.47\textwidth]{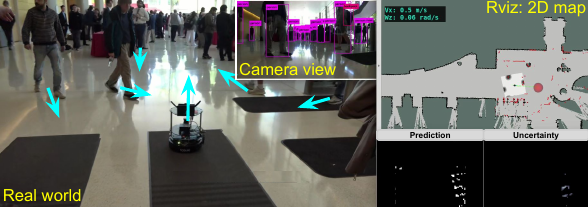}
            \label{fig:hallway_drl_vo_soscope_scope4}
    }%
    \subfloat[t+3]{
            \centering
            \includegraphics[width=0.47\textwidth]{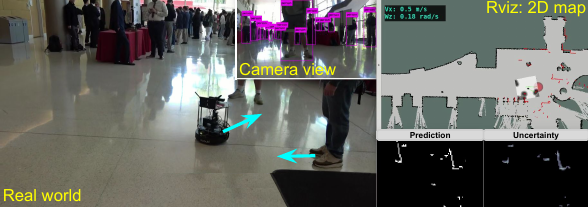}
            \label{fig:hallway_drl_vo_soscope_scope5}
    }%

    \caption{Robot deployed with DRL-VO/SO-SCOPE/PU reactions to moving pedestrians in the indoor hallway with the high crowd density at different times.
            }
    \label{fig:hallway_drl_vo}
\end{figure*}

\subsection{Hardware Results}
\label{subsubsec:hardware_results}
Besides simulated experiments, we also conduct real-world experiments to demonstrate the applicability of our DWA/SCOPE/PU and DRL-VO/SO-SCOPE/PU policies.
Both control policies use the \texttt{amcl} ROS package to provide the robot localization in known maps.  
Note that our DRL-VO/SO-SCOPE/PU control policy requires additional running of three deep learning networks (\ie YOLOv3~\cite{redmon2018yolov3}, SO-SCOPE, and DRL-VO~\cite{xie2023drlvo}) and one optimization-based multiple hypothesis tracker (MHT)~\cite{yoon2018multiple} to implement corresponding pedestrian detection, environment OGM prediction, navigation control, and pedestrian tracking.
From the attached Multimedia and \cref{fig:hallway_dwa}, we can see how our robot deployed with DWA/SCOPE/PU can actively avoid collisions with walking students crossing the hallway, safely avoid static students standing, and reach predefined goals by following predictive uncertainty-aware nominal paths, traveling a total length of \unit[76.10]{m} and an average speed of \unit[0.42]{m/s}.
It demonstrates the real-world effectiveness of our proposed SCOPE predictor and DWA/SCOPE/PU planner. 

In addition, from the attached Multimedia and \cref{fig:hallway_drl_vo}, we can see that even with three learning-based blocks, our robot deployed with  DRL-VO/SO-SCOPE/PU is still able to quickly and actively avoid collisions with walking students crossing the hallway and reach predefined goals by following predictive uncertainty-ware nominal paths, traveling a total length of \unit[86.41]{m} and an average speed of \unit[0.47]{m/s}.
It demonstrates our software-optimized SO-SCOPE predictor is hardware friendly and our software-optimized predictive uncertainty-aware navigation framework can be combined with different high computational load learning-based algorithms.

\section{Conclusion}
\label{sec:conclusion}
In this article, we propose a family of hardware-friendly stochastic occupancy grid map prediction algorithms (\ie SCOPE++, SCOPE, and SO-SCOPE) that provides mobile robots with the ability to accurately and robustly predict the future states of complex, dynamic, human-occupied environments.
Specifically, we first propose two VAE-based stochastic predictors (\ie SCOPE++ and SCOPE) that exploit information from robot motion, object motion, and static scene geometry, and use a VAE-based network to fuse this useful information into the latent space and predict the distribution of future environment states.
We then perform software optimization on them using knowledge distillation and uncertainty quantification operations to propose a hardware-friendly SO-SCOPE, which significantly improves the inference speed, addresses their sampling-based memory-intensive nature, and extends their practical applications in resource-constrained robots.
Furthermore, we propose a novel predictive uncertainty-aware navigation framework by combining these proposed stochastic predictors with the costmap-based ROS navigation stack to improve the performance of current state-of-the-art model-based and learning-based control policies for safe navigation in crowded dynamic scenes.

We demonstrate that our proposed SCOPE family achieves smaller absolute error, higher structure similarity, and higher tracking accuracy than the other state-of-the-art image-based predictors on three different simulated and real-world datasets collected by three different robot models.
It also provides a range of plausible and diverse future prediction states for complex stochastic environments.
We further demonstrate through operational analysis and experiments on embedded computing devices that our proposed software-optimized SO-SCOPE prediction engine is much faster and more memory-efficient, allowing it to provide real-time inference with other resource-intensive algorithms in resource-constrained robots. 
Lastly, we demonstrate through simulated and hardware experiments that we can easily integrate the predicted maps into the existing navigation framework, all of which benefit from leveraging the prediction and uncertainty information of our SCOPE family.
In summary, our proposed SCOPE predictors are hardware-friendly and improve mobile robots' ability to safely navigate through crowded dynamic environments.


\section*{Acknowledgment}
\label{sec:acknowledgment}
This research includes calculations carried out on HPC resources supported in part by the National Science Foundation through major research instrumentation grant number 1625061 and by the US Army Research Laboratory under contract number W911NF-16-2-0189.

\bibliographystyle{IEEEtran}
\bibliography{bib/Dames,bib/refs}
\vspace{-15mm}

\begin{IEEEbiography}[{\includegraphics[width=1in,height=1.25in,clip,keepaspectratio]{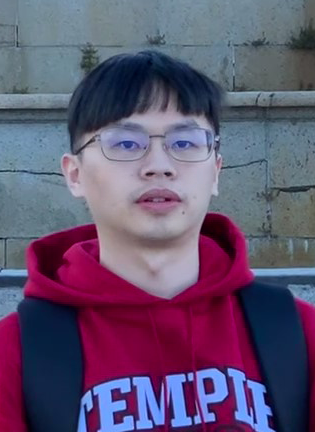}}]{Zhanteng Xie}
    received the B.Eng. degree in electronic information engineering from Zhengzhou University, Zhengzhou, China, in 2015, the M.Eng. degree in information and communication engineering from the Harbin Institute of Technology, Harbin, China, in 2018, and the Ph.D. degree in mechanical engineering, Temple University, Philadelphia, USA. From 2018 to 2019, he was a research assistant in the Department of Electrical and Electronic Engineering at the Southern University of Science and Technology, Shenzhen, China. 
    
    His research interests lie at the intersection of robotics and machine learning, with a focus on environment perception, environment prediction, and autonomous robot navigation in crowded dynamic scenes.
\end{IEEEbiography}
\vspace{-15mm}

\begin{IEEEbiography}[{\includegraphics[width=1in,height=1.25in,clip,keepaspectratio]{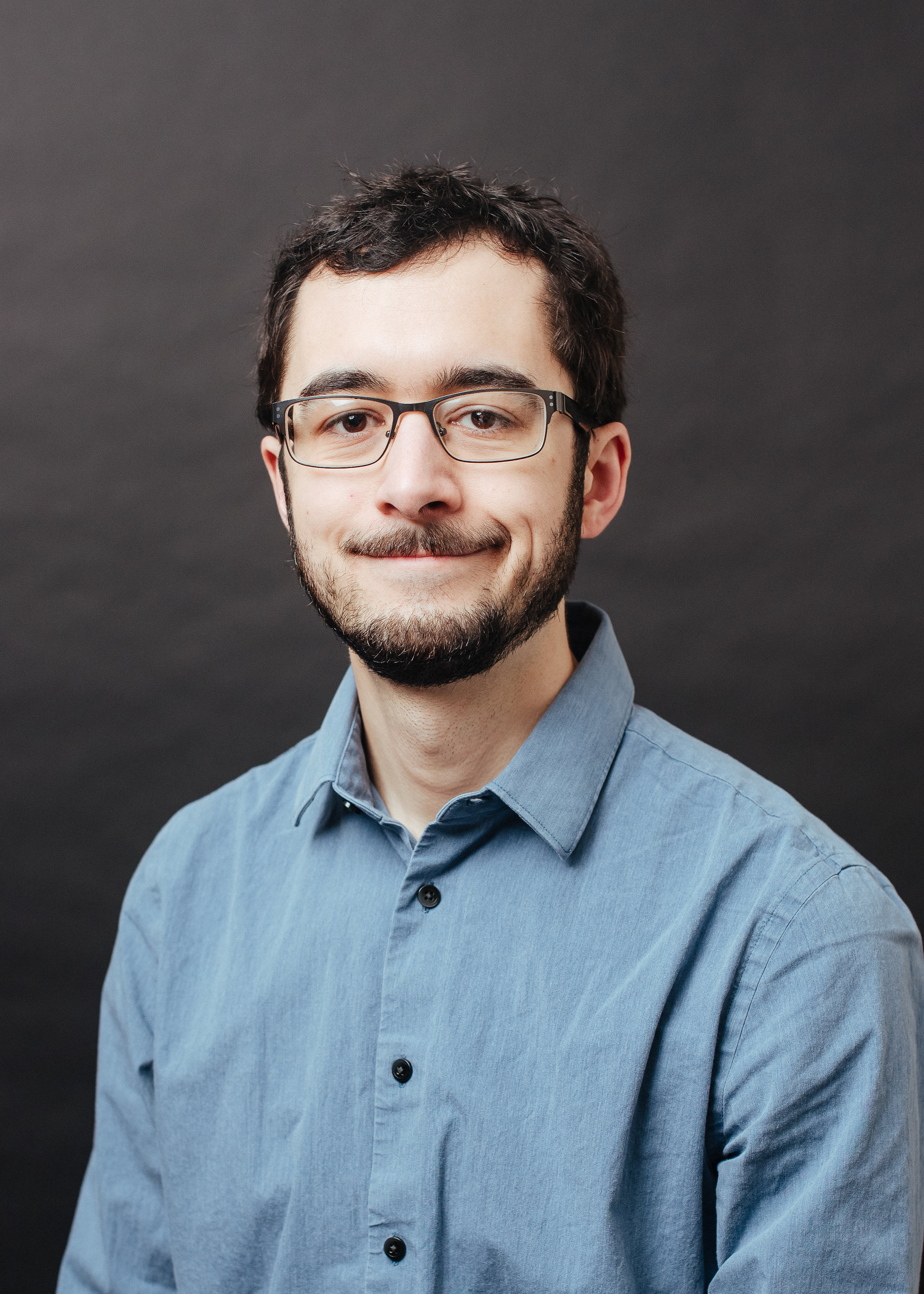}}]{Philip Dames}
    received both his B.S. (summa cum laude) and M.S. degrees in mechanical engineering from Northwestern University, Evanston, IL, USA, in 2010 and his Ph.D. degree in mechanical engineering and applied mechanics from the University of Pennsylvania, Philadelphia, PA, USA, in 2015. 
    
    From 2015 to 2016, he was a Postdoctoral Researcher in Electrical and Systems Engineering at the University of Pennsylvania. Since 2016, he has been at Temple University, Philadelphia, PA, USA, where he is currently an Associate Professor of Mechanical Engineering and directs the Temple Robotics and Artificial Intelligence Lab (TRAIL). His research aims to improve robots’ ability to operate in complex, real-world environments to address societal needs.

    He is a member of IEEE and is the recipient of an NSF CAREER award.
\end{IEEEbiography}


\end{document}